\documentclass{article}

\PassOptionsToPackage{authoryear, compress}{natbib}

\usepackage[final]{neurips_2019}

\usepackage{placeins}
\usepackage[utf8]{inputenc} 
\usepackage[T1]{fontenc}    
\usepackage{todonotes}
\usepackage[colorlinks,citecolor=blue]{hyperref}       
\usepackage{url}            
\usepackage{booktabs}       
\usepackage{amsmath,amsthm,amsfonts,amssymb}
\usepackage{nicefrac}       
\usepackage{microtype}      
\usepackage{wrapfig}
\usepackage{cancel}
\usepackage{placeins}

\usepackage{lipsum}
\usepackage{color}

\usepackage{multirow}
\usepackage{array}
\newcolumntype{P}[1]{>{\centering\arraybackslash}p{#1}}

\usepackage{float}
\usepackage{graphicx,subcaption,rotating,float}
\usepackage[labelfont=bf]{caption}

\DeclareMathOperator*{\argmax}{arg\,max}

\usepackage[ruled,vlined]{algorithm2e} 
\usepackage{setspace}
\SetKwComment{Comment}{$\triangleright$\ }{}
\SetCommentSty{itshape}

\usepackage{cleveref}
\crefformat{section}{§#2#1#3}

\makeatletter
\def\thm@space@setup{%
  \thm@preskip=2pt plus 1pt minus 1pt
  \thm@postskip=\thm@preskip 
}
\makeatother

\theoremstyle{definition}
\newtheorem{definition}{Definition}[section]

\title{Boltzmann Exploration Expectation--Maximisation}

\author{%
  Mathias~Edman and Neil Dhir \\
  Kamin AI \\
  Linn\'{e}gatan 89C, 115 23, Stockholm, Sweden \\
  \texttt{\{mathias,neil\}@kamin.ai}\\
}

\begin{document}
\maketitle
\begin{abstract}
We present a general method for fitting finite mixture models (FMM). Learning in a mixture model consists of finding the most likely cluster assignment for each data-point, as well as finding the parameters of the clusters themselves. In many mixture models this is difficult with current learning methods, where the most common approach is to employ monotone learning algorithms e.g. the conventional expectation-maximisation algorithm. While effective, the success of any monotone algorithm is crucially dependant on good parameter initialisation, where a common choice is $K$-means initialisation, commonly employed for Gaussian mixture models.

For other types of mixture models the path to good initialisation parameters is often unclear and may require a problem specific solution. To this end, we propose a general heuristic learning algorithm that utilises Boltzmann exploration to assign each observation to a specific base distribution within the mixture model, which we call Boltzmann exploration expectation-maximisation (BEEM). With BEEM\footnote{Code: \url{https://github.com/kaminAI/beem}}, hard assignments allow straight forward parameter learning for each base distribution by conditioning only on its assigned observations. Consequently it can be applied to mixtures of any base distribution where single component parameter learning is tractable. The stochastic learning procedure is able to escape local optima and is thus insensitive to parameter initialisation. We show competitive performance on a number of synthetic benchmark cases as well as on real-world datasets.
\end{abstract}

\section{Introduction}
\label{sec:intro}

Mixtures of generative models are ubiquitous in the machine learning, statistical and data-science literature. Prominent examples include the Gaussian mixture model (GMM) \citep{lindsay1995mixture,Bishop2006,Murphy2012}, the mixture of hidden Markov models (MHMM) \citep{rabiner1989tutorial, krogh1994hidden, smyth1997clustering,couvreur1996hidden, helske2017mixture, subakan2014spectral, panuccio2002hidden}, mixture of experts (MoE) \citep{rasmussen2000infinite}, mixture of Gaussian processes (MGP) \citep{lazaro2012overlapping, rasmussen2000infinite, yuan2009variational}, as well as more recent additions such as the mixture of generative adversarial networks (MGAN) \citep{mukherjee2018clustergan}. Application domains are plentiful an include e.g. multi-object target-tracking \citep{lazaro2012overlapping}; speaker-identification \citep{reynolds1995robust} as well as document clustering \citep{blei2003latent} to mention but a few.

When fitting FMMs, we are required to find the parameters that maximise the likelihood of the observations.
To fit a GMM for example, it is common to adopt the expectation-maximisation (EM) algorithm
\citep{dempster1977maximum}. \citet{blomer2013simple} explain that the EM algorithm alternates between
computing a lower bound of the log-likelihood and improving the current model w.r.t. this lower bound. In concluding the
learning, the procedure converges to a particular stationary point on the likelihood function. However; the likelihood
function is generally non-convex, possesses many stationary points, includes small local maxima and minima as well as
saddle points. Importantly, the convergence of the EM algorithm to either type of point, \emph{depends crucially on the
initialisation} \citep{blomer2013simple}. 


Initialisation sensitivity is not particular to GMMs however, the same problems arises for e.g. MHMMs as well, and indeed many of the other aforementioned mixture models. In particular those which rely on EM as the main driver of inference. To deal with this problem we propose a general method which carries across mixture model modalities (i.e. can be used for clustering sequences as well as objects) and, compared to many of the current methods in the respective mixture model domains, is comparatively simple. Our EM strategy uses the principle of maximum entropy, where our main contributions are:
\begin{enumerate}
    \itemsep0em
    \item \emph{Boltzmann exploration} (also known as softmax action selection) \citep[\S 2.8]{sutton2018reinforcement}, is used to select a cluster (arm) for each observation. This is done under the standard reinforcement learning aegis, where each arm is selected with a probability proportional to its average reward \citep{kuleshov2014algorithms}. 
    \item Conventional EM \citep{dempster1977maximum} seeks out zero-gradients in the log-likelihood landscape, under maximum-likelihood estimation, and is thus prone to get stuck. To counter this we follow the Boltzmann update with an exploration step which seeks to avoid local maxima, minima and saddle-points.
    \item BEEM is applied to GMMs, MHMMs and MGPs, and therein demonstrates competitive performance on synthetic and real datasets.
\end{enumerate}

The paper is organised as follows: we review relevant background material in \cref{sec:back}, in \cref{sec:method} we present our method, in \cref{sec:related_work} we discuss related work and in \cref{sec:experiments} we conduct experiments on synthetic and real data. A discussion and conclusion closes the paper in \cref{sec:concl_disc}.

\section{Background}
\label{sec:back}

By adopting the mixture modelling paradigm, complex distributions can be constructed from $K$ simpler components, even describing arbitrary densities when $K \rightarrow \infty$. In this paper we turn to mixture models, not to model complex distributions, but to find modalities in our observed data by the way of clustering.  We adopt the view of mixtures presented by \citet[\S9]{Bishop2006}, where discrete latent variables are interpreted as defining assignments of observations to particular mixture components. This interpretation allows us to employ methodology and nomenclature often found in the space of reinforcement learning.

\theoremstyle{definition}
\begin{definition}{(Mixture model).}
    \label{def:mm}
    The mixture model is so prescribed because we are mixing together $K$ base distributions \citep{Murphy2012} as a
    linear combination, yielding a joint likelihood of the form:
    \begin{equation}
        \label{eq:mix_model}
        p(x_1,\ldots,x_n,\ldots,x_N,z_1,\ldots,z_n,\ldots,z_N \mid \zeta, \eta) = \prod_{n=1}^N  p ( x_n \mid z_n, \zeta) \cdot p(z_n\mid\eta)
    \end{equation}
    where $p(z_n\mid \eta)$ is the mixture weight for \emph{latent} mixture index (component) $z_n \in \{1,\ldots,K\}$ with parameters $\eta$. Further, $p(x_n \mid z_n, \zeta)$ is the likelihood model for observation $x_n$
    with parameters $\zeta$ \citep{Barber2012}. The most likely assignment of observations is given by
    \begin{equation}
        \label{eq:big_argmax}
        \argmax_{z_1,\ldots,z_n,\ldots,z_N} p(z_1,\ldots,z_n,\ldots,z_N \mid x_1,\ldots,x_n,\ldots,x_N).
    \end{equation}
    \Cref{eq:mix_model} shows that we can factorise the joint, consequently the expression in \cref{eq:big_argmax} is equivalent \citep[\S 20.1]{Barber2012} to computing $\argmax_{z_n} p(z_n \mid x_n) \ \forall n \in \{1,\ldots,N\}$. Because cluster assignments are typically \emph{a priori} unknown (in unsupervised learning), they need to be located via inference. Consequently, the optimal cluster assignment
    \begin{equation}
        \label{eq:learn_mm}
        p(x_1,\ldots,x_n,\ldots,x_N \mid \zeta , \eta ) = 
        \argmax_{\zeta, \eta} \prod_{n=1}^N \left [\sum_{z_n}  p ( x_n \mid z_n, \zeta) \cdot p(z_n\mid\eta) \right]
    \end{equation}
     can be inferred via an optimisation procedure. Numerically this can be achieved using e.g. a gradient based approach, or, when the cluster indices are explicitly latent, we can apply an EM algorithm \citep{Barber2012,Murphy2012}.
\end{definition}

As noted in \cref{sec:intro}; the linchpin of our method is the Boltzmann distribution or the the softmax function which we use to explore possible clusters for any given observation. The Boltzmann distribution is a measure which gives the probability that a system will be in a certain state as a function of that state's energy and the temperature of the system. In our setting we replace `state' with \emph{cluster}. Further, for completeness and clarity, multi-armed bandits are defined in \cref{def:bandit}.



\begin{definition}{(Multi-armed bandit).}
    \label{def:bandit}
    A multi-armed bandit is a rather unusual slot machine which instead of having one arm, has $K$. A player, always seeking to maximise her reward, will select actions i.e. pull one of the $K$ levers, and receive rewards --different rewards-- depending on the reward distribution of that arm. Through repeated action-selection she will maximise her winnings by concentrating her actions on the best levers \citep{sutton2018reinforcement}.
\end{definition}

\begin{definition}{(Boltzmann exploration).}
    \label{def:softmax_def}
    Softmax action selection methods are based on Luce’s axiom of choice \citep{luce2012individual} and selects each arm (see \cref{def:bandit}) with a probability given by the Boltzmann distribution, that is proportional to its average reward \citep{kuleshov2014algorithms}. As such, one version of  Softmax action selection, selects arm $z_n$ on the $n^{th}$ play, using the Boltzmann distribution
    \begin{equation}
        \label{eq:arm_selection}
        p(z_n \mid \tau ) = \frac{\exp  (\widehat{\mu}_{z_n} / \tau )}{\sum_{n=1}^N \exp  (\widehat{\mu}_{z_n} / \tau )}
    \end{equation}
    where $\widehat{\mu}_{z_n}$ is the empirical average of the rewards\footnote{Note that in thermodynamics it is typical to negate the exponent's argument, as this corresponds to a lowest energy state having the highest probability.} obtained from arm $z_n$ up until round $n$. High temperatures $\tau$ cause the actions to be all (nearly) equiprobable, whereas low temperatures cause a greater difference in selection probability for actions that differ in their value estimates. In the limit as $\tau \rightarrow \infty$, Softmax action selection becomes the same as greedy action selection \citep{kuleshov2014algorithms,sutton2018reinforcement}.
\end{definition}

\section{Boltzmann exploration expectation-maximisation}
\label{sec:method}

Using the material provided in \cref{sec:back} herein we describe a method that uses a modified version of Softmax action selection, to learn mixtures of generative models with a view of overcoming complex initialisation schmes by way of exploration-exploitation.


As alluded to in the above paragraph, we seek a principled way to maximise the expression in \cref{eq:learn_mm} w.r.t the model parameters. In \cref{sec:related_work} and \cref{sec:EM_init_appendix} we discuss state-of-the-art methods for initialising such optimisation procedures. What follows is a novel and simple method for achieving this goal using a modified gradient bandit algorithm for learning mixture-model component assignments. Unlike the EM procedure for e.g. GMMs or MHMMs, we do not seek to compute the responsibility weighted parameter updates, but rather assign each sample to a mixture component straight away and subsequently update each component as a likelihood model conditioned on the assigned samples. We will annotate the method exposition analogous to the EM procedure to promote ease of comparison. 

\paragraph{Component assignment (expectation step)}
\label{sec:component_assignment}

In the classical EM setting, the expectation step (E-step) for mixture models \citep[\S 20.2]{Barber2012} utilises the update 
\begin{equation}
    \label{eq:E_update}
    p^{\text{new}} (z_n = k \mid x_n, \zeta_{t},\eta_{t}) \propto  p ( x_n \mid z_n, \zeta_{t-1}) \cdot p(z_n\mid\eta_{t-1})
\end{equation}
where the right-hand side is also called the \emph{responsibility} \citep[\S 11.4]{Murphy2012}, which cluster $z_n = k$ takes for observation $x_n$. We use index $t$ to keep track of the iteration count. The E-step takes the following simple form, which is the same for any mixture model:
\begin{equation}
    r_{n,z_n} \triangleq p^{\text{new}} (z_n = k \mid x_n,\zeta_t,\eta_t ) =  \frac{p ( x_n \mid z_n, \zeta_{t-1}) \cdot p(z_n\mid\eta_{t-1})}{\sum_{z_n} p ( x_n \mid z_n, \zeta_{t-1}) \cdot p(z_n\mid\eta_{t-1})}.
    \label{eq:Estep}
\end{equation}
Our first contribution modifies the standard E-step in \cref{eq:Estep} by fixing uniform mixing weights $p(z_n \mid \eta) = \frac{1}{K}$ and instead determines the cluster responsibility via the Boltzmann distribution vis-\`{a}-vis \cref{def:softmax_def}, by substituting \cref{eq:E_update} into \cref{eq:arm_selection}:
\begin{align}
    \label{eq:marauderEstep}
    r_{n,z_n} &=
    \frac{ \exp \left [ \log p ( x_n \mid z_n, \zeta_{t-1}) / \tau +
     \log p(z_n\mid\eta_{t-1}) / \tau \right ]
    }{\sum_{z_n} \exp \left [\log p ( x_n \mid z_n, \zeta_{t-1}) / \tau + \log p(z_n\mid\eta_{t-1}) / \tau \right ]} \\
    &= 
    \frac{ p ( x_n \mid z_n, \zeta_{t-1})^{\frac{1}{\tau}} 
      p(z_n\mid\eta_{t-1})^{\frac{1}{\tau}}
    }{\sum_{z_n} p ( x_n \mid z_n, \zeta_{t-1})^{\frac{1}{\tau}} 
      p(z_n\mid\eta_{t-1})^{\frac{1}{\tau}}
      } \\
    &= 
    \frac{ p ( x_n \mid z_n, \zeta_{t-1})^{\frac{1}{\tau}} 
      \cancelto{(1/K)^{\frac{1}{\tau}}}{p(z_n\mid\eta_{t-1})^{\frac{1}{\tau}}}
      }
    {\sum_{z_n} p ( x_n \mid z_n, \zeta_{t-1})^{\frac{1}{\tau}} 
      \cancelto{(1/K)^{\frac{1}{\tau}}}{p(z_n\mid\eta_{t-1})^{\frac{1}{\tau}}}}
\end{align}
Having asserted uniform mixing weights, we introduce a \emph{modified responsibility}:
\begin{equation}
    \label{eq:BEEM_final}
    r_{n,z_n}^{\prime} \triangleq p^{\text{new}} (z_n = k \mid x_n,\zeta_t) = 
    \frac{ p ( x_n \mid z_n, \zeta_{t-1})^{\frac{1}{\tau}}}
    {\sum_{z_n} p ( x_n \mid z_n, \zeta_{t-1})^{\frac{1}{\tau}}}
\end{equation}
where we have taken the logarithm of the exponential function's arguments. 

To understand the rationale behind the algorithm design, it is helpful to view the task of finding optimal cluster assignments\footnote{Assignment rewards are not independent of past decisions, thus not conforming to the Markov property.} roughly as a  bandit problem. We pose the problem as follows; an agent is faced with the task of assigning samples to the components in such a way that maximises the accumulated maximum data likelihood $\sum_t^{T}\sum_{n}^{N}\underset{z_n}{\max} \{ p(x_n \mid z_n, \zeta_{t,z_n}) \}$ at horizon $T$. Since each assignment is made in the context of $x_n$ and $\zeta_{t}$ are updated at each iteration, our task is similar to that of a non stationary contextual bandit. While contextual bandits explicitly model the expected reward for each action, we simply let $ p(x_n \mid z_n, \zeta_{t-1,z_n})$, be a noisy estimate of the reward at iteration $t$. In order to introduce an exploration mechanism, we sample a hard component assignment from the responsibility distribution, rather than computing responsibility weighted parameter updates as in the EM algorithm. The motivation for fixing uniform mixing weights is \emph{to avoid the probability of assignments being amplified for large clusters}, thus resulting in smaller clusters being assigned less samples and eventually vanishing. Consequently, we have constructed an algorithm which at each iteration learns a better similarity metric and tries to group similar samples together. We concede that our problem does not qualify under the definition of a bandit, however it is a apt analogy. 


Now, the EM procedure monotonically increases the observed data log-likelihood until it reaches a local maxima, minima or saddle-point \citep[\S 11.4.7]{Murphy2012}. But there is nothing which prevents the method from getting stuck at either of these points, as EM only seeks out zero-gradients in the log-likelihood landscape. Consequently, to prevent this, we introduce the aforementioned exploration mechanism as part of the component assignment. A new component assignment for observation $x_n$, is sampled as:
\begin{equation}
\label{eq:sample_component}
    z_n \sim \text{Categorical} \left (K,[r_{n,1},\ldots, r_{n,K}] \right)
\end{equation}
using the responsibilities found in \cref{eq:BEEM_final} when $z_n \in \{1,\ldots,K\}$. One full pass of this update results in a new or the same assignment for each observation. Where the temperature in \cref{eq:BEEM_final} is allowed to decrease with each EM update, to reduce the level of exploration. To achieve this we introduce an exponential cooling schedule $g(\tau,t,\alpha) \triangleq \tau \alpha ^{t-1}$ which takes as input the current temperature $\tau$, the step count $t$ and decay factor $0< \alpha< 1$. The full BEEM procedure is presented in \cref{alg:marauder}.



Because of the way BEEM is constructed it means that once the component assignment has been updated, parameter inference continues as usual, depending on the mixture base-class we are currently working with. This means that if we are considering e.g. a GMM then the parameter update is achieved through the maximisation step of the usual EM algorithm -- no additional change of this step is required, and it is fully congruent on the component assignment. Or if we are using a mixture of HMMs, then we would employ the Baum-Welch algorithm \citep{rabiner1989tutorial} for the parameter update.

\subsection{Relationship with standard expectation-maximisation}

There are two major differences between standard EM and BEEM. First, BEEM is not a monotonically increasing algorithm due to its exploration mechanism. Consequently it is less sensitive to parameter initialisation. Certainly, one may argue that EM could emulate the same property by allowing multiple random initialisation. However, foregoing a regret analysis one would expect that approach to require far more iterations than BEEM (we empirically demonstrate this in \cref{sec:experiments}). Secondly, rather than directly maximising the complete data likelihood conditioned on all mixture components, BEEM separately maximises the data likelihood for samples conditioned on the assigned component.

\begin{algorithm*}[ht!]
    \setstretch{1.2}
    \caption{Boltzmann exploration expectation-maximisation (BEEM) \label{alg:marauder}}
    \DontPrintSemicolon
    \SetKwInOut{Input}{Input}
    \SetKwInOut{Output}{Output}
    \Input{$\mathcal{X} = \{x_1,\ldots, x_N \}$, $\{\zeta_1,\ldots,\zeta_K\}$, $\varepsilon, \tau, \alpha, g(\tau,t,\alpha)$}
    \BlankLine
        Let $V \in \mathbb{R}_+^{N \times K}$ \hfill \Comment{Value matrix}
        $\{\mathcal{S}_1,\ldots,\mathcal{S}_k,\ldots,\mathcal{S}_K\} \leftarrow$ random split  $\{\mathcal{X}\}$ \hfill \Comment{Assign random observation subsets}  
        
        \For{$k \in \{1,\ldots,K\}$}{
                
                Condition model $\zeta_k$ on observations in $\mathcal{S}_k$\hfill \Comment{Initialise model parameters}}

        $t=0$ \hfill \Comment{Iteration index}
        \While{$\zeta_{t} - \zeta_{t-1} > \varepsilon $}{

            $\{ \mathcal{S}_k \leftarrow \emptyset \mid k = 1,\ldots,K  \}$ \hfill
            \Comment{Reset observation subsets}
            
            \For{$n \in \{1,\ldots,N \}$}{
            
             \For{$k \in \{1,\ldots,K\}$}{
                    $v_{n,k} \leftarrow \log p(x_n \mid  \zeta_{z_n})$ \hfill \Comment{Update value matrix $V$ using the base model PDF}
                    }

            $\{r'_{n,1},\ldots,r'_{n,K} \} \sim p(z_n \mid v_{n,1}, \ldots, v_{n,K}, \tau )$ \hfill \Comment{Sample modified responsibility --see \cref{eq:BEEM_final}}
            
            $z_n \sim \text{Categorical}(K,[r'_{n,1},\ldots,r'_{n,K}])$  \hfill \Comment{Sample observation assignment -- see \cref{eq:sample_component}}
            $\mathcal{S}_{z_n} \leftarrow x_{n}$ \hfill \Comment{Assign observation $x_n$ to subset $\mathcal{S}_{z_n}$ indexed by $z_n$}
            
            }
            
            \For{$k \in \{1,\ldots,K\}$}{
                Condition model $\zeta_k$ on observations in $\mathcal{S}_k$\ \hfill \Comment{M-Step}
                   
            }

            $\tau \leftarrow g(\tau,t,\alpha)$ \hfill \Comment{Update temperature}
            
            $t \leftarrow t + 1$\; 
        }
    \BlankLine
    \Output{Parameters $\{\zeta_1,\ldots,\zeta_K\}$ which in the limit segments the observations into $K$ bins.}
\end{algorithm*}

\section{Related work}
\label{sec:related_work}

BEEM is a type of deterministic annealing (DA) algorithm \citep[\S 11.4.9]{Murphy2012}. The field of DA is an active area of research, and has been for the past few decades. There are many prominent pieces of literature, worth discussing but herein we only discuss the most relevant and compare and contrast to one of these methods in \cref{sec:experiments}. Additional discussion can be found in \cref{sec:EM_init_appendix}.

The goal of FMM fitting is to find the global maximum in the MLE landscape. This goal, when using EM, is heavily dependent upon the initialisation. For relevant work regarding initialisation schemes which do \emph{not} modify the core EM procedure \citep{dempster1977maximum}, see \cref{sec:EM_init_appendix}. One method which does modulate the core EM mechanism is the algorithm by \citet{ueda1998deterministic}, where the authors introduced the deterministic annealing EM (DAEM) algorithm which uses the principle of maximum entropy. The DAEM algorithm is perhaps the one method which bears the greatest resemblance to ours. Interestingly, the DAEM algorithm was published in the same year that \citet{rose1998deterministic} published his tutorial paper on DA methods in clustering and other application domains. The difference between BEEM and DAEM is that DAEM does not sample observation assignments, nor do they partition the observation space into $K$ clusters, into which observations are hard assigned (which BEEM does, based on the sampled cluster index). Instead they assign observations to clusters by using a weighted responsibility calculation (see the DAEM algorithm in \citep{ueda1998deterministic}) -- i.e. theirs relies on a more `fuzzy' clustering approach \citep{rose1998deterministic} since they assign membership based on the measured probabilities. Another prominent method is the split-and-merge EM (SMEM) algorithm \citep{ueda2000smem}. The SMEM procedure overcomes more complex local maxima problems which DAEM struggles with. \citet{ueda2000smem} empirically show that SMEM is marginally better than DAEM in their toy-data experiments, but at a higher computational cost.

\subsection{Bayesian treatment}
\label{sec:bayesian}

At first pass, the exposition which begins with \cref{eq:marauderEstep}, looks like prior specification. But this interpretation would be incorrect. A Bayesian treatment is indeed attractive and is something that we have explored for GMM learning. However, posterior sampling for FMM learning is fundamentally different to Thompson sampling for bandits. 

In a stationary bandit setting, each update of the posterior over reward distribution parameters, leads to an improved estimate of the true parameters (conditioned on us knowing which action and observed reward originates from). But in the cluster-learning setting the ground-truth cluster-responsibility for samples is unknown. Consequently, when sampling observations are used for the next posterior update, each base distribution is \emph{not} guaranteed to be updated with observations from only one cluster. This in turns means that the learned FMM is not guaranteed to converge to the data-generating distribution. The posterior for each base distribution will be informed by the entire history of sampled observations and so, with each iteration, becomes increasingly difficult to modulate. This means that a bad initialisation will penalise the whole learning process, where the prior will reinforce erroneous cluster allocations. To overcome this a successful implementation would likely include some form of discount mechanism analogous to Discounted Thompson Sampling \citep{raj2017taming} for restless bandits. We see this as a interesting direction for future work, that would offer a more principled probabilistic motivation. But as it stands, BEEM should be employed as a heuristic for mixtures where posterior updates are intractable or impractical.
\section{Experiments}
\label{sec:experiments}
In this section we investigate a number of different FMMs, applied to real and synthetic datasets. We compare inference in these models using BEEM alongside other state-of-the-art methods as well as standard approaches. Reported metrics are; normalized mutual information (NMI), adjusted Rand index (ARI) \citep{Hubert1985}, clustering purity (ACC), and homogenity score \citep{rosenberg2007v} (Homo). The BEEM hyper parameters are set to $\tau = 1.5$ and $\alpha = 0.97$ for all experiments. The optimisation was terminated after the maximum complete data log-likelihood $\sum_{n}^{N}\underset{z_n}{\max} \{ p(x_n \mid z_n, \zeta_{t,z_n}) \}$ failed to improve for 10 EM-steps. Unless otherwise stated, each experiment to was repeated 100 times. Error bounds are found within brackets in each results table. Finally, we consider two different types of mixing weights for BEEM and two initialisation methods for GMMs described in \cref{tab:beem_weights} and \cref{tab:beem_inits} respectively.

\begin{table}[!htbp]
    \centering
    \caption{Initialisation methods.}
    \label{tab:beem_inits}
    \def\arraystretch{1.1}%
    \begin{tabular}[t]{lcl}
        \toprule 
        Initialisation & Key & Description \\
        \midrule
        Random & A & \begin{tabular}{@{}l@{}}  The initial cluster means are randomly \\ drawn from the set of observations.\end{tabular}   \\
        $K$-means & B & \begin{tabular}{@{}l@{}}  The initial cluster means are computed \\  via the $K$-means algorithm. \end{tabular}  \\
        \bottomrule 
    \end{tabular}
\end{table}

\begin{table}[!htbp]
    \centering
    \caption{Mixing weight types.}
    \label{tab:beem_weights}
    \def\arraystretch{1.1}%
    \begin{tabular}[t]{lcl}
        \toprule 
        Mixing weight type & Key & Description \\
        \midrule
        Fixed uniform & I &  \begin{tabular}{@{}l@{}} Mixing weights are fixed to be uniform\\  throughout the optimisation.                                 \end{tabular} \\
        Standard weights & II & \begin{tabular}{@{}l@{}}  Mixing weights are updated at every M-step \\  as in the standard EM algorithm.                         \end{tabular} \\
        \bottomrule 
    \end{tabular}
\end{table}

Note that we will only entertain different mixing-weights types in \cref{sec:GMM_experiments}, to explore performance as a consequence of this factor. From thereon, unless otherwise stated, BEEM mixing weights are \emph{only} fixed uniform. 

\subsection{Gaussian mixture models}
\label{sec:GMM_experiments}

In this section we use GMMs with BEEM on a number of synthetic and real datasets. We include a comparison to EM with 100 re-initialisation (EM 100\footnote{\label{em100} Results reported for EM 100 is that of the solution with highest complete data log-likelihood.}) to gauge the difficulty of each dataset. For the rainbow dataset (described overleaf), we conduct a larger comparison study, including recently published methods which includes: power $k-$means clustering (power) \citep{pmlr-v97-xu19a}; hierarchical density-based spatial clustering of applications with noise (HDBSCAN) \citep{mcinnes2017hdbscan} and affinity propagation (AP) \citep{frey2006mixture}. For power $k$-means, we explored $s_0 \in \{-1,-3\}$ as it is not clear from the original publication what constitutes an appropriate initial value (note that $s_0$ can take any value in the domain $-\infty <s_0 <0$). We got the best results when $s_0=-1$ as the original paper suggested (see  \S 4 and tables 1 and 2 of the original publication).

\subsubsection{Unbalanced square simulation}\label{par:square} 
\begin{figure*}[ht]
    \centering
    \begin{subfigure}[t]{0.48\textwidth}
        \centering
        \includegraphics[width = \columnwidth]{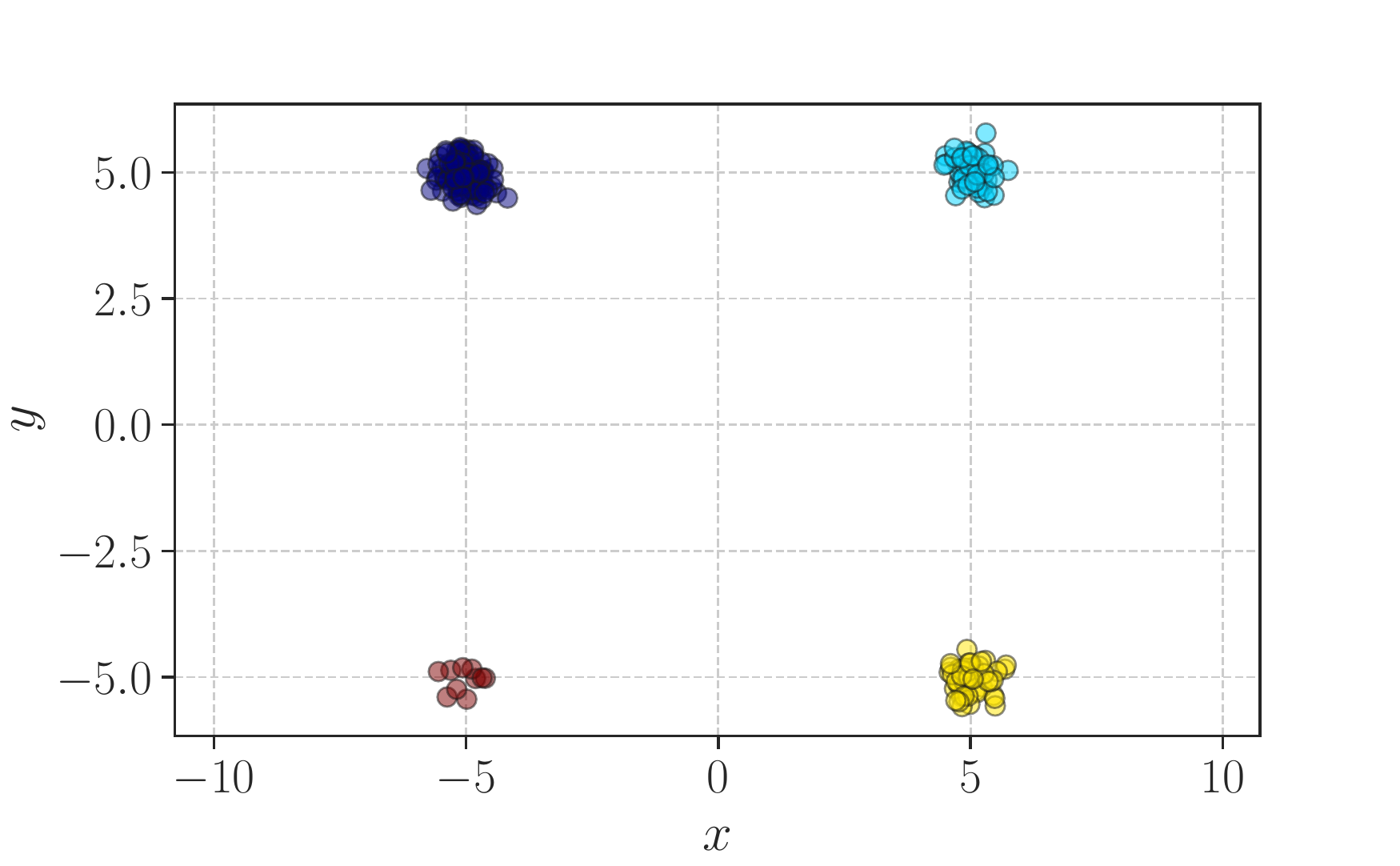}
        \caption{Unbalanced square simulation. From the top left corner in the clockwise direction the clusters contain 100, 50, 50, and 10 samples.\label{fig:unbalanced_fig}}
    \end{subfigure}%
    \hfill
    \begin{subfigure}[t]{0.48\textwidth}
        \centering
        \includegraphics[width = \columnwidth]{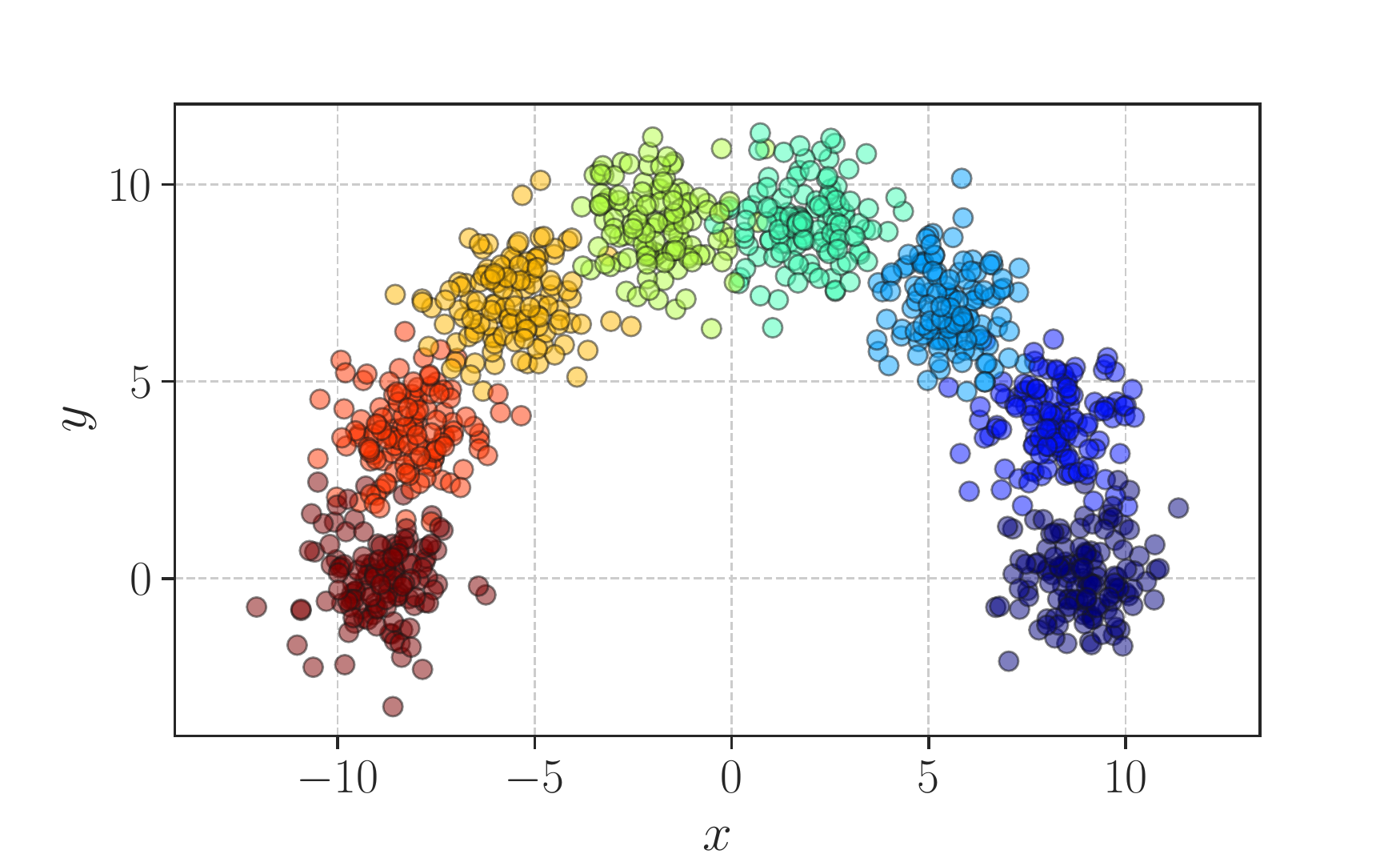}
        \caption{Rainbow simulation. Showing eight clusters evenly distributed along a half circle in the first and second quadrant.\label{fig:rainbow_fig}}
    \end{subfigure}
    \caption{Visualisation of simulated clustering tasks employed for the GMM experiments. The colour of each sample indicates which cluster generated that data point.\label{fig:gmm_mixture_figs}}
\end{figure*}

A clustering task consisting of four two dimensional normal distribution located at the corners of a square with side length 10, centred at the origin -- see \cref{fig:unbalanced_fig}. All distributions have covariance matrix  $\Sigma =0.3 \mathbf{I}_2$.  To create an unbalanced clustering scenario the number of samples drawn from each cluster is set to $[100,50,50,10]$. As can be seen in \cref{fig:unbalanced_fig}, this constitutes a simple clustering task for the human eye. However it is used here to demonstrate the vanishing cluster artefact that arises from including mixing weights in the BEEM algorithm.  Results are shown in \cref{tab:square}.

\FloatBarrier

\begin{table}[!htbp]
    \centering
    \caption{Unbalanced square simulation results  -- mean and (std.).}
    \label{tab:square}
    \def\arraystretch{1.1}%
    \scalebox{0.95}{
    \begin{tabular}[t]{lccccccc}
        \toprule 
        Method         & Init. & Weight & ACC & Homo & NMI & ARI & EM steps \\
        \midrule
        BEEM & A & I &  $0.99 \ (0.02)$ & $ 0.96\ (0.05)$ & $0.96\ (0.07)$ & $0.94\ (0.11)$ & $39.84$ $(12.48)$\\
        BEEM & A & II & $0.85 \ (0.11)$ & $ 0.76\ (0.18)$ & $0.85\ (0.12)$ & $0.72\ (0.22)$  & $-$\\
        EM 100 & A & $-$ & $\mathbf{1.00 \ (0.00)}$ & $\mathbf{1.00 \ (0.00)}$ & $\mathbf{1.00 \ (0.00)}$ & $\mathbf{1.00 \ (0.00)}$ & $20.89$ $3.98$ \\
        EM & A & $-$ & $0.86 \ (0.11)$ & $ 0.76\ (0.18)$ & $0.86\ (0.11)$ & $0.75\ (0.21)$ & $12.02 \ (9.45)$\\
        DAEM & A & $-$ & $0.85$ $(0.12)$ & $0.76$ $(0.19)$ & $0.84\ (0.12)$ & $0.73$ $(0.22)$ & $-$\\
        \bottomrule 
    \end{tabular}
    }
\end{table}

As seen in \cref{tab:square}, BEEM is competitive on this synthetic clustering task. In the  `method' header in \cref{tab:square} we have included different initialisation methods for  the conventional EM algorithm. We see that BEEM is competitive with both conventional initialisations. Further, DAEM proved robust in \cref{tab:square}. Comparing the number of EM steps\footnote{The EM steps for DAEM were omitted given that the result varies greatly with the choice of hyper parameters.}, BEEM maintains competitive efficiency to EM, especially considering that the BEEM M-step only requires each base model to be updated w.r.t. a subset of the dataset while for EM each base model is updated on the complete dataset. Our contribution outperforms EM 100, both in terms of EM steps and cluster metrics, indicating that the proposed exploration mechanism is sound. Finally we see that BEEM with mixing weights performs worse than the suggested uniform mixing weights. This is due to vanishing cluster artefact that arises when the probability of assignment to large clusters is amplified by the mixing weights.

\subsubsection{Balanced square simulation}\label{par:balanced_square} 
We continue with a simpler clustering task still, wherein each cluster is of the same size. Here we aim to draw the reader's attention to the discussion conducted in \cref{sec:bayesian} w.r.t. to the Bayesian treatment of EM. Alas, this clustering task consists of four two dimensional Gaussian distribution located at the corners of a square with side length 10, centred at the origin -- see \cref{fig:unbalanced_fig}. All distributions have covariance matrix  $\Sigma =0.3 \mathbf{I}_2$. Results are shown in \cref{tab:balanced_square}.

\begin{table}[!htbp]
    \centering
    \caption{Balanced square results $[K=4$, $x_n \in \mathbb{R}^2, N=200]$ -- mean and (std.).}
    \label{tab:balanced_square}
    \def\arraystretch{1.05}%
    \footnotesize
    \scalebox{1.}{
    \begin{tabular}[t]{lcccccc}
        \toprule 
        Method & Initialisation & Weight & ACC & Homo & NMI & ARI \\
        \midrule
        BEEM  & A & I &  $\mathbf{0.91 \ (0.12)}$ & $\mathbf{0.90 \ (0.13)}$ & $\mathbf{0.93 \ (0.10)}$ & $\mathbf{0.87\ (0.18)}$ \\
        BEEM & A & II & $0.75 \ (0.09)$ & $ 0.74\ (0.09)$ & $0.85\ (0.07)$ & $0.71\ (0.09)$ \\
        \bottomrule 
    \end{tabular}
    }
\end{table}

\begin{figure*}[ht]
    \captionsetup[subfigure]{aboveskip=-1pt,belowskip=1pt}
    \centering
    \begin{subfigure}[t]{0.49\textwidth}
        \centering
        \includegraphics[width = \columnwidth]{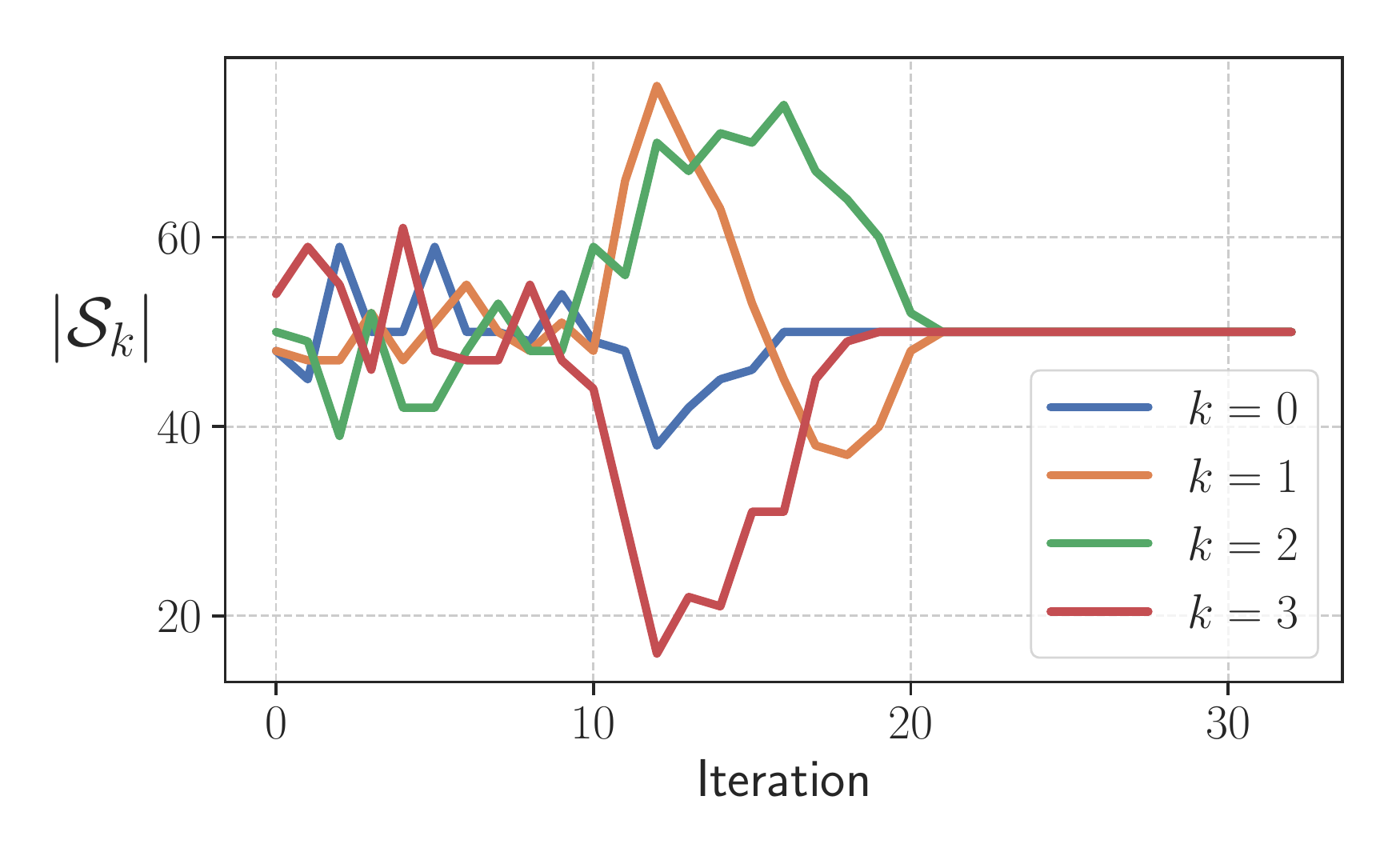}
        \caption{Uniform.\label{fig:beem_uniform}}
    \end{subfigure}%
    \hfill
    \begin{subfigure}[t]{0.49\textwidth}
        \centering
        \includegraphics[width = \columnwidth]{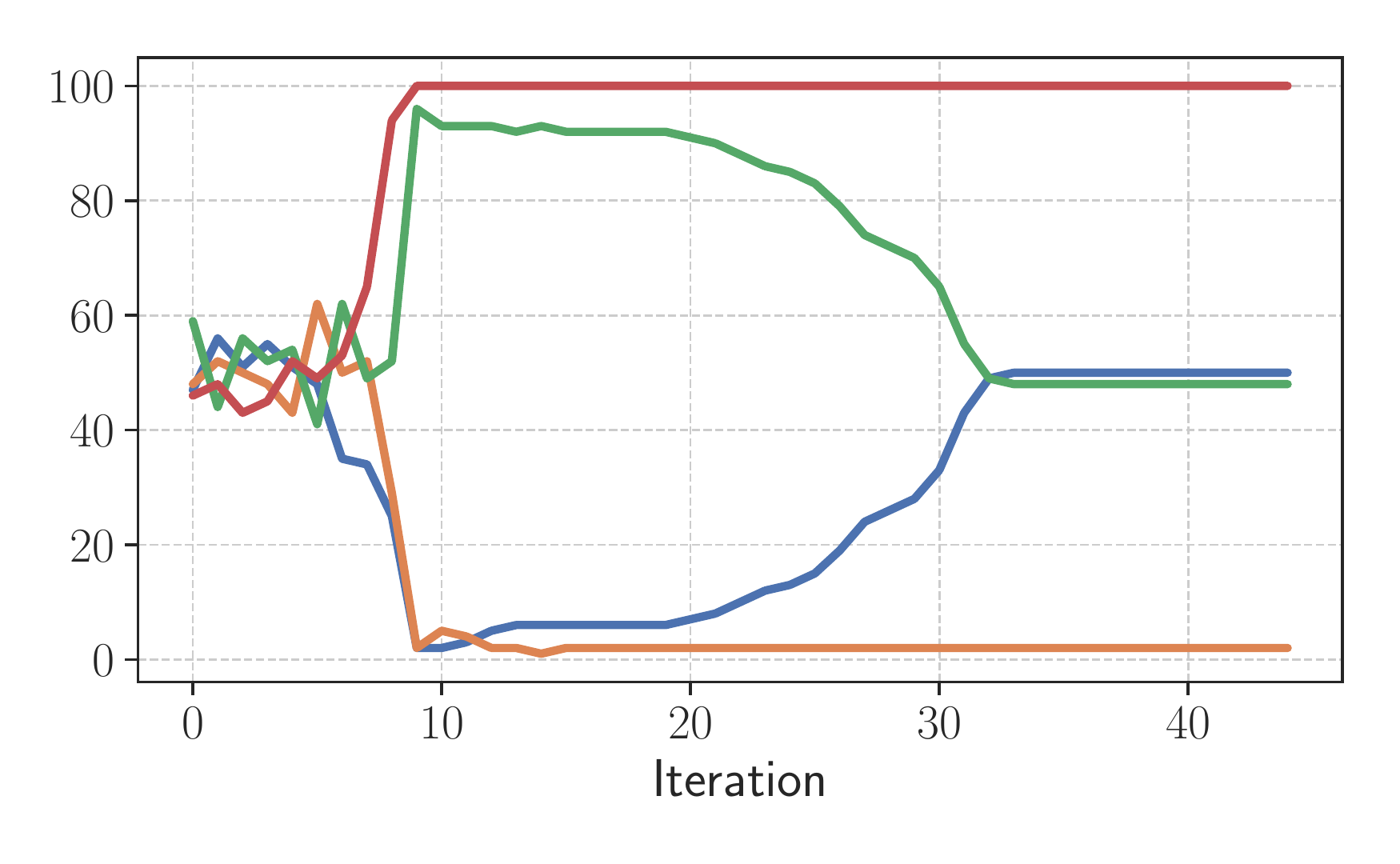}
        \caption{Mixing.\label{fig:beem_mixing}}
    \end{subfigure}%
    \caption{Visualisation of number of samples $|\mathcal{S}_k|$ allocated to each base distribution during learning on the \emph{balanced} square dataset. For completeness, BEEM seeks to allocate an equal number of samples to each of the four clusters. \label{fig:beem_compare}}
\end{figure*}

In this simple example we see that employing a Bayesian treatment of the cluster assignments, is not necessarily a recipe for success. Recall that the posterior for each base distribution will be informed by the entire history of sampled observations and so, with each iteration, becomes increasingly difficult to modulate. This means that a bad initialisation will penalise the whole learning process. Hence, even in this trivial synthetic example, does not suggest that employing a prior is, in fact, beneficial. Indeed, consider the visualised learning processes in \cref{fig:beem_compare}.

In this instantiation of the learning process (repeated 100 times to yield the results found in \cref{tab:balanced_square}) we observe that the learning process has indeed been penalised by a poor intialisation, which it cannot readily escape from. This phenomena is seen too for BEEM with mixing weights in \cref{fig:beem_mixing} but not with main version of BEEM in \cref{fig:beem_uniform}.  

\subsubsection{Rainbow simulation.}\label{par:rainbow} The rainbow clustering task consists of eight, partially overlapping, two-dimensional Gaussian distributions with identity covariance matrices and distribution means along a half circle with radius $r=9$ -- see \cref{fig:rainbow_fig}. The means are homogeneously distributed by letting the angle, $\boldsymbol \omega = \{0,\frac{\pi}{8},\frac{2\pi}{8},\ldots,\pi \}$ and calculating their $x$ and $y$ mean location as, $\mu_i = [\text{Re} \{ r\exp(j\omega_i) \},\text{Im} \{ r\exp(j\omega_i)\}]$, where $j$ is the complex unit. Samples are drawn with uniform probability from each distribution for a total of 1000 samples. In this experiment we also explore BEEM with $K$-means intialisation -- to mimic the most common initialisation scheme for standard EM. Results are shown in \cref{tab:rainbow}.
\begin{table}[ht]
    \centering
    \caption{Rainbow simulation results -- mean and (std.).}
    \label{tab:rainbow}
    \setlength{\tabcolsep}{3pt}
    \def\arraystretch{1.1}%
    \scalebox{0.9}{
    \begin{tabular}[t]{lcccccc}
        \toprule 
        Method   & Initialisation      & ACC & Homo & NMI & ARI & EM steps \\
        \midrule
        EM & B &  $0.93 \ (0.05)$ & $ 0.89\ (0.04)$ & $0.89\ (0.04)$ & $0.87\ (0.08)$ & $3.60$ $(1.83)$  \\  
        BEEM & B & $\mathbf{0.96 \ (0.00)}$ & $\mathbf{ 0.91\ (0.01)}$ & $\mathbf{ 0.91\ (0.01)}$ & $\mathbf{ 0.91\ (0.01)}$ & $11.00$ $(0.00)$ \\
        \midrule
        EM 100 & A & $0.49$ $(0.14)$ & $0.51$ $(0.14)$ & $0.62$ $(0.10)$ & $0.41$ $(0.14)$ & $13.09$ $(16.92)$  \\
        EM & A &  $0.43 \ (0.05)$ & $ 0.46\ (0.06)$ & $0.61\ (0.05)$ & $0.33\ (0.07)$ & $2.68$ $(4.85)$\\  
        BEEM & A & $0.93 \ (0.05)$ & $ 0.89\ (0.04)$ & $0.89\ (0.04)$ & $0.87\ (0.06)$ & $76.88$ $(17.18)$\\
        \midrule
        DAEM & A & $0.75$ $(0.08)$ & $0.75$ $(0.06)$ & $0.79$ $(0.06)$ & $0.66$ $(0.09)$ & $-$ \\
        AP & $-$ & $0.93\ (0.01)$ & $\mathbf{0.91\ (0.01)}$& $0.78\ (0.03)$& $0.59\ (0.06)$ & $-$\\
        HDBSCAN & $-$ & $0.70\ (0.09)$& $0.66\ (0.07)$& $0.67\ (0.04)$& $0.46\ (0.08)$ & $-$\\
        power $(s_0 = -1)$ & $-$ & $0.94\ (0.05)$& $0.90\ (0.04)$& $0.90\ (0.03)$& $0.88\ (0.07)$ & $-$ \\
        power $(s_0 = -3)$ & $-$ & $0.91\ (0.06)$& $0.88\ (0.05)$& $0.88\ (0.04)$& $0.84\ (0.09)$ & $-$ \\
        power $(s_0 = -9)$ & $-$ & $0.87\ (0.07)$& $0.85\ (0.05)$& $0.85\ (0.05)$& $0.78\ (0.09)$ & $-$ \\
        power $(s_0 = -18)$ & $-$ & $0.85 \ (0.07)$& $0.83\ (0.05)$& $0.84\ (0.05)$& $0.75\ (0.09)$ & $-$ \\
        \bottomrule 
    \end{tabular}
    }
\end{table}
\FloatBarrier

Unlike the squared simulations, this synthetic dataset is far more complex. The Bayesian treatment of the problem reflects this imminently, by performing poorly, despite being provided with the closed-form posterior update. Most likely this owes to the nature of the data, where clusters are overlapping, thus making assignment difficult (particularly if the assignment scheme relies on complete assignment histories). BEEM with $K$-means initialisation performs the best, as too does BEEM with uniform assignment weights. Vanilla EM with $K$-means comes close to BEEM as does power, where the former does so in far fewer EM steps.

\subsubsection{Fisher's Iris dataset}\label{par:iris}  
The dataset \citep{fisher1936use,Dua:2019} contains three classes of 50 instances each, where each class refers to a type of Iris flower. Each sample is represented by an attribute vector (sepal length, sepal width, petal length, petal width). Results are shown in \cref{tab:iris}.

\begin{table}[!htbp]
    \centering
    \caption{Iris dataset results $[K=3$, $x_n \in \mathbb{R}^4, N=150]$  -- mean and (std.).}
    \label{tab:iris}
    \def\arraystretch{1.1}%
    \scalebox{0.9}{
    \begin{tabular}[t]{lcccccc}
        \toprule 
        Method   & Initialisation      & ACC & Homo & NMI & ARI & EM steps\\
        \midrule
        EM  & B  &  $\mathbf{0.97 \ (0.02)}$ & $\mathbf{0.90\ (0.03)}$ & $\mathbf{0.90\ (0.03)}$  & $\mathbf{0.91\ (0.04)}$ & $16.86$ $(0.98)$\\
        BEEM & B &  $\mathbf{0.97\ (0.03)}$ & $\mathbf{0.90\ (0.01)}$ & $\mathbf{0.90\ (0.02)}$ & $0.90\ (0.04)$ & $27.67$ $(5.53)$\\ 
        \midrule
        EM 100 & A & $0.81$ $(0.13)$ & $0.72$ $(0.14)$ & $0.76$ $(0.10)$ & $0.68$ $(0.16)$ & $29.68$ $(8.21)$ \\
        EM & A &  $0.76\ (0.05)$ & $0.61\ (0.06)$ & $0.62\ (0.06)$ & $0.55\ (0.06)$ & $22.31$ $(6.05)$\\
        BEEM & A &  $0.87\ (0.07) $ & $0.72\ (0.10)$ & $ 0.73\ (0.10)$ & $0.69\ (0.11)$ & $52.09$ $(0.09)$         \\
        \midrule
        DAEM & A & $0.78$ $(0.02)$ & $0.61$ $(0.01)$ & $0.62$  $(0.01)$ & $0.55$ $(0.01)$ & $-$ \\
        \bottomrule 
    \end{tabular}
    }
\end{table}

\subsubsection{Discussion}

The results reported in tables \cref{tab:square}, \cref{tab:balanced_square}, \cref{tab:rainbow} and \cref{tab:iris} show that BEEM is competitive on all selected datasets. Further, under the `method' header in each table we have also included different initialisation methods for each conventional EM algorithm. In addition DAEM proved robust in \cref{tab:square} but less so in the other two experiments. Comparing the number of EM steps\footnote{The EM steps for DAEM were omitted given that the result varies greatly with the choice of hyper parameters.}, BEEM maintains competitive efficiency to EM, especially considering that the BEEM M-step only requires each base model to be updated w.r.t. a subset of the dataset while for EM each base model is updated on the complete dataset. Our contribution outperforms EM 100, both in terms of EM steps and cluster metrics, indicating that the proposed exploration mechanism is sound.
Finally we see that BEEM with mixing weights performs worse than the suggested uniform mixing weights on the unbalanced square simulations. This is due to vanishing cluster artefact that arises when the probability of assignment to large clusters is amplified by the mixing weights.  

\subsection{Mixtures of hidden Markov models}

Graduating from GMM parameter estimation we consider the more challenging mixtures of hidden Markov models. As revealed by the name, the base distribution of a MHMM (i.e. a single HMM) is in itself a latent variable model and as such, can be considered as a mixture model. A MHMM is simply a special case of an ordinary HMM in which the transition matrix is restricted in order to partition the state space, e.g if we consider a mixture of $K$ HMMs with transition matrices $A_k$, then the resulting MHMM transition matrix is a block diagonal matrix with elements, $\{ A_1, \ldots ,A_K \}$. 

A common approach for HMM parameter initialisation is to set uniform transition probabilities and set the means and covariances using the $K$-means algorithm. Though sensible for a single HMM, it does not generalise to the mixture case. Consider for example the task of segregating sequences drawn from HMMs with identical emission distributions but distinct dynamics i.e. transition matrices. To extend $K$-means initialisation procedure, \citet{smyth1997clustering} suggests that the inter-sequence similarity can be estimated by fitting a HMM to each individual sequence in the data set, and constructing a $N\times N$ similarity matrix by computing the log-likelihood of each of the $N$ sequences w.r.t. the individual models. $K$-means clustering then proceeds on the similarity matrix such that $K$ sets of parameters can be initialised from the resulting segmentation.

In this section we compare BEEM parameter estimation to the Baum-Welch algorithm with random and \citet{smyth1997clustering} initialisation.

\subsubsection{Random HMM simulation}
 \label{par:randomhmm}
We simulate a clustering task by at each iteration initialising \emph{three} distinct HMMs $(K=3)$ with random transition matrices and initial distributions, each with \emph{four} hidden states. The state distributions are fixed and equal for all clusters with state means $\mu = [-2,-1,0,1]$ and standard deviation $\sigma = 0.1$. At every iteration, 20 new sequences are drawn, with random sequence length $L\sim \mathcal{U}(a,b)$, from each HMM. The experiment is run for two different sequence length settings, $(a,b) = (5,10)$ and $(a,b) = (20,50)$.

\begin{table}[!htbp]
    \centering
    \caption{Random HMM simulation \cref{par:randomhmm} results  -- mean and (std.).}
    \label{tab:blob}
    \def\arraystretch{1.1}%
    \begin{tabular}[t]{c|lccccc}
        \toprule 
        $ L $ & Method & Initialisation & ACC & Homo & NMI & ARI \\
        \midrule
        \multirow{3}{*}{$5-10$} 
        & EM & A   & $\textbf{0.51 (0.07)}$ & $\textbf{0.13 (0.07)}$ & $\textbf{0.13 (0.08)}$  & $\textbf{0.09 (0.07)}$       \\ 
        & EM & \citet{smyth1997clustering} & $0.47$ $(0.07)$ & $0.09$ $ (0.07)$ & $0.11$ $ (0.08)$  & $0.05$ $(0.07)$       \\ 
        & BEEM & A  & $0.49$ $(0.06)$ & $0.09$ $(0.06)$ & $0.10$ $(0.7)$ & $0.07$ $(0.07)$       \\ 
         \midrule
        \multirow{3}{*}{$20-50$} 
        & EM & A & $0.80 \ (0.15)$ & $0.64 \ (0.22)$ & $0.67 \ (0.20)$ & $0.60 \ (0.24)$ \\
        & EM & \citet{smyth1997clustering} & $0.71 $ $(0.13)$ & $0.51 (0.18)$ & $0.56$ $(0.18)$ &
        $0.49$ $(0.20)$\\ 
        & BEEM & A & $\textbf{0.87 (0.11)}$ & $\textbf{0.68 (0.19)}$ & $\textbf{0.69 (0.19)}$ & $\textbf{0.68 \ (0.21)}$ \\ 
        \bottomrule 
    \end{tabular}
\end{table}

\subsubsection{Character trajectories} This data set \citep{Dua:2019} contains three dimensional ($x$, $y$, pressure) pen tip trajectories for handwritten characters. We utilise a subset of the characters to create two clustering tasks. Separation of characters `A' and `B' $(K=2)$, and separation of characters `A' to `E' $(K=5)$. The average sequence length is 117. For each character the following instance-count holds: \{`A': 171, `B': 141, `C': 142, `D': 157, `E': 186\}.
\label{par:character}

\begin{table}[!htbp]
    \centering
    \caption{Character trajectories data set \cref{par:character} -- best result for each algorithm after three initialisations.}
    \label{tab:hmm_char}
    \def\arraystretch{1.1}%
    \begin{tabular}[t]{c|lccccc}
        \toprule 
        $K$ & Method & Initialisation & ACC & Homo & NMI & ARI \\
        \midrule
        \multirow{3}{*}{2 (`A'-`B')} 
        & EM & A & $1.0$ & $1.0$ & $1.0$  & $1.0$       \\ 
        & EM & \citet{smyth1997clustering} & $1.0$ & $1.0$ & $1.0$  & $1.0$       \\ 
        & BEEM & A \& I & $1.0$ & $1.0$ & $1.0$  & $1.0$       \\ 
        \midrule
        \multirow{3}{*}{5 (`A'-`E')} 
        & EM & A & $0.96$ & $0.90$ & $0.91$  & $0.90$       \\ 
        & EM & \citet{smyth1997clustering} & $0.96$ & $0.90$ & $0.91$  & $0.89$       \\ 
        & BEEM & A & $\textbf{0.98}$ & $\textbf{0.95}$ & $\textbf{0.95}$  & $\textbf{0.96}$ \\ 
        \bottomrule 
    \end{tabular}
\end{table}

\subsubsection{Discussion}

The results in \cref{tab:hmm_char} and \cref{tab:blob} demonstrate the utility of using BEEM with a mixture of HMMS. In \cref{tab:hmm_char} we see that conventional methods as well as \citet{smyth1997clustering}'s sequence clustering algorithm, perform well on that dataset when the number of clusters is small. However, when the number of clusters increases their performance decreases, as too does BEEM, but not as much. Similarly in \cref{tab:blob} BEEM outperforms both methods, when it comes clustering sequences of random length $20-50$. When the sequences are of length $5-10$, there is no distinguishable difference between the methods, which is to be expected given that this is a very difficult clustering task.

\subsection{Mixtures of (overlapping) Gaussian processes}
\label{sec:gp_exp}
\begin{wrapfigure}[13]{r}{0.4\textwidth}
    \vspace{-1.3em}
    \centering
    \includegraphics[width=\linewidth]{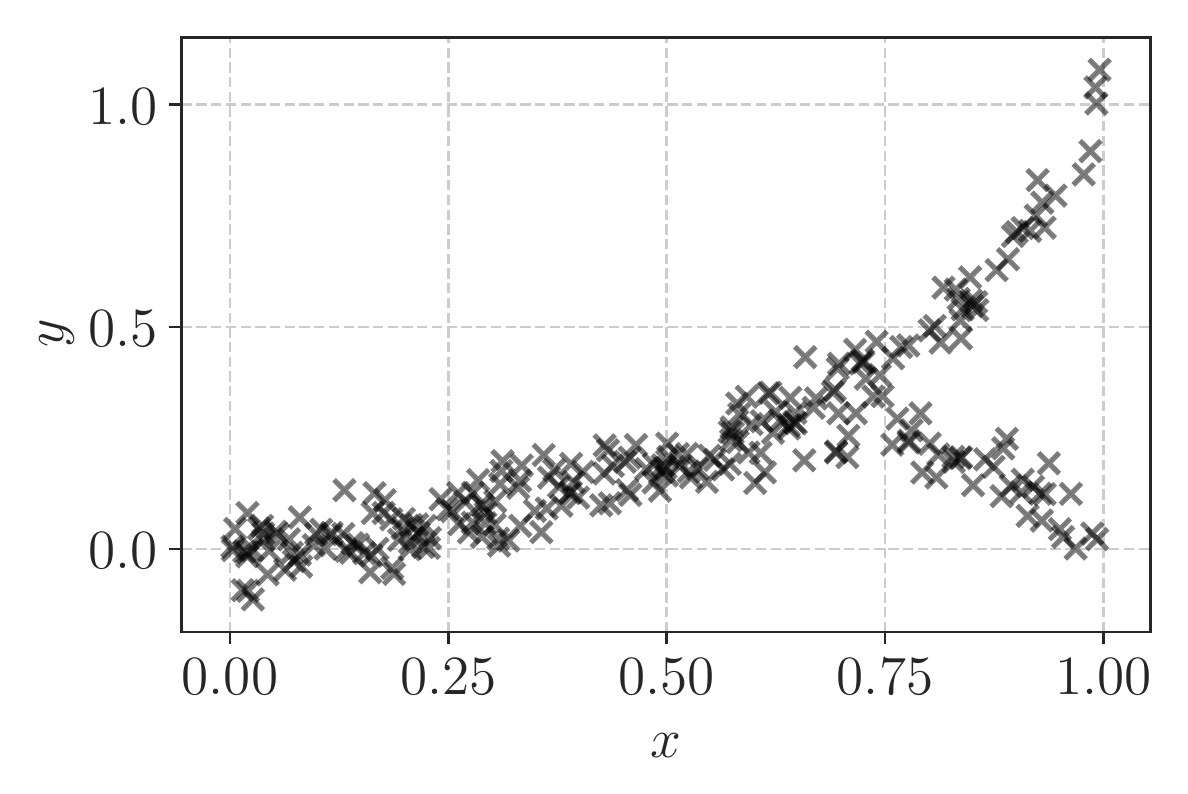}
    \vspace{-2em}
    \caption{Data association finds the latent model(s) responsible for generating the observations.\label{fig:data_assoc}}
\end{wrapfigure}

\begin{figure*}[h]
    \vspace{-0.5em}
    \centering
    \begin{subfigure}[t]{0.49\textwidth}
        \centering
        \includegraphics[width = \columnwidth]{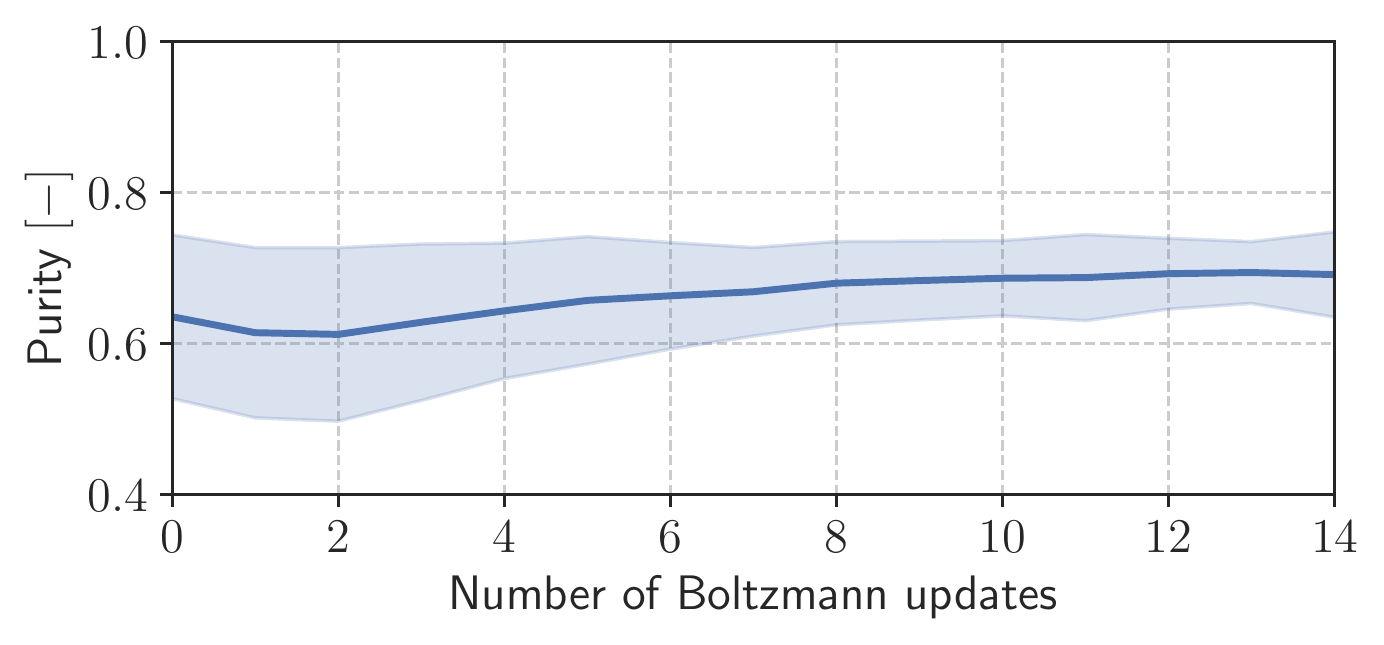}
        \caption{Measured cluster purity for MGP w. BEEM.}
    \end{subfigure}%
    ~ 
    \begin{subfigure}[t]{0.49\textwidth}
        \centering
        \includegraphics[width = \columnwidth]{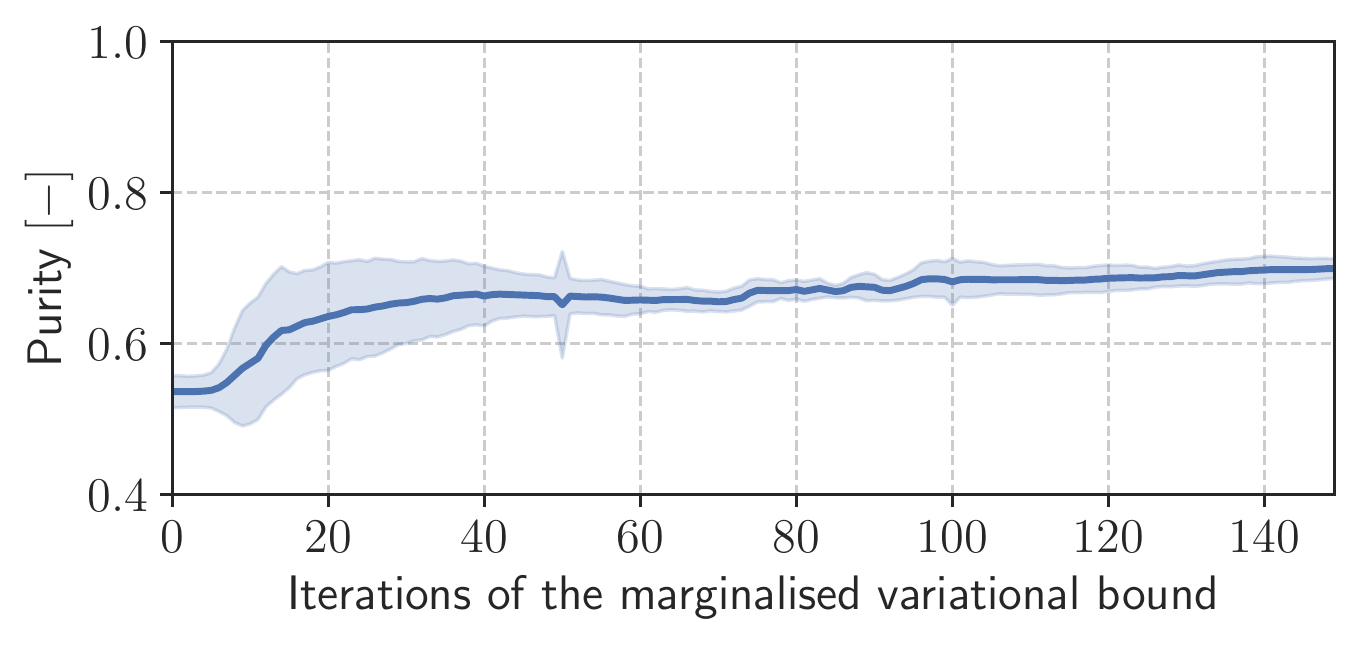}
        \caption{Measured cluster purity for the OMGP.}
    \end{subfigure}

    \caption{Measured purity (ACC) for each method (higher is better). The thick line in each plot shows the mean trend $\mu \pm 2\sigma$. Note that the horizontal axis is \emph{not} the same for both methods.\label{fig:main_body_gp_results}}
\end{figure*}

In this section we investigate a simple \emph{data association} (DA) problem. Data association seeks to map a source-model to each observation in the dataset, see \cref{fig:data_assoc}, in which the true number of clusters is $K=2$ and $N=250$. A prominent model which deals in DA, is the overlapping mixtures of Gaussian processes (OMGP) model by \citet{lazaro2012overlapping}.
\citet{lazaro2012overlapping} use a ``non-standard variational Bayesian algorithm to efficiently recover sample labels and learn the hyperparameters''. To this we compare a simple mixture of Gaussian processes (MGP) using BEEM inference. As with the other mixtures we use a off-the-shelf inference for learning the likelihood model, which in the case of the GP means we minimise negative log marginal-likelihood of each base-model w.r.t. the hyperparameters and noise level. We use the same radial basis function kernel \citep{Rasmussen:2005:GPM:1162254} for each model, with noise variance set to $0.01$ and $K=2$ for both models. Each experiment, for each model, was repeated 50 times. The OMGP's marginalised variational bound was minimised over 150 iterations for each experiment and the MGPs with BEEM, used 15 Boltzmann updates and ten GP updates each, per component, at each exploration. The cluster purity results are shown in \cref{fig:main_body_gp_results}.

The results in \cref{fig:main_body_gp_results} demonstrate that a standard mixture of GPs using BEEM inference, is capable of achieving similar purity results as the more complex OMGP model, when applied to the toy dataset in \cref{fig:data_assoc}. For additional results and experiments see \cref{appendix:gps}.

\section{Conclusion and discussion}
\label{sec:concl_disc}

We have presented the Boltzmann exploration expectation-maximisation algorithm for maximum likelihood estimation. BEEM overcomes many of the problem associated with the conventional EM algorithm \citep{dempster1977maximum}, as well as more modern alternatives such as DAEM and EM with $K$-means initialisation. While an elegant theoretical motivation for BEEM has been omitted from this paper, we have empirically shown that it manages to effectively search the parameter-space while avoiding local maximums. In addition to strong performance the algorithm is very simple to implement for finite mixtures of any base distribution, given that their parameters are updated independently at each iteration. This means, as demonstrated, that model-specific parameter inference can be used, off-the-shelf, whilst the component assignment can be effectively done using BEEM, without any major change in the overall model fitting procedure.

Finally, in this work we control the exploration level via exponentially decreasing temperature. The algorithm is insensitive to hyper-parameter settings given that the same parameter set was used successfully for all experiments. However, it is possible that the used parameter set caused excessive exploration leading to an unnecessarily large number of EM-steps. Further directions include exploring the possibility of a self regulating exploration mechanism to improve algorithm efficiency.

\bibliography{references}
\bibliographystyle{icml2019}
\appendix
\section{Additional results for mixture of Gaussian processes experiment}
\label{appendix:gps}
\FloatBarrier

\begin{figure*}[ht]
    \centering
    \begin{subfigure}[t]{0.49\textwidth}
        \centering
        \includegraphics[width = \columnwidth]{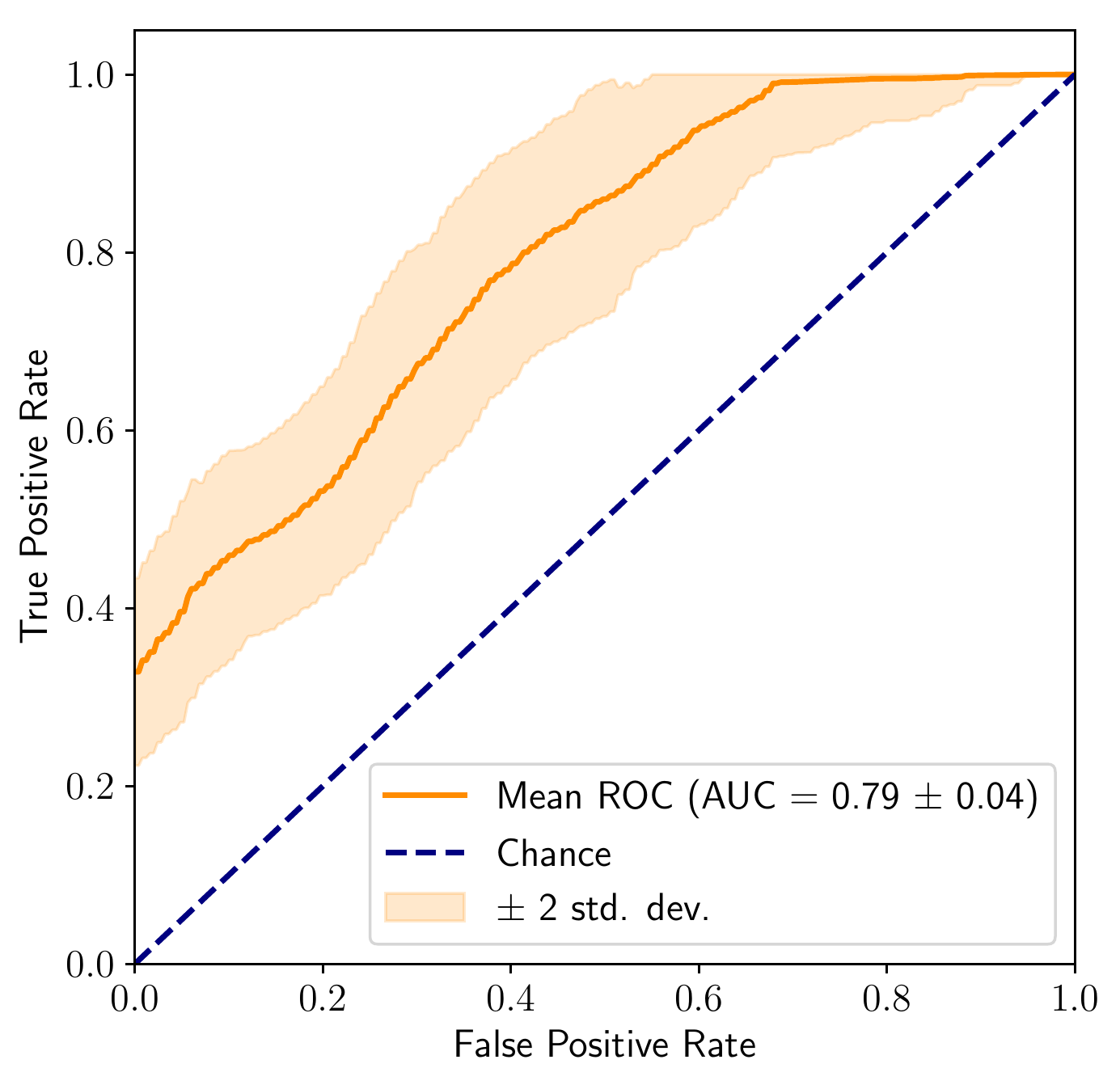}
        \caption{MGP w. BEEM.\label{fig:roc_beem}}
    \end{subfigure}%
    ~ 
    \begin{subfigure}[t]{0.49\textwidth}
        \centering
        \includegraphics[width = \columnwidth]{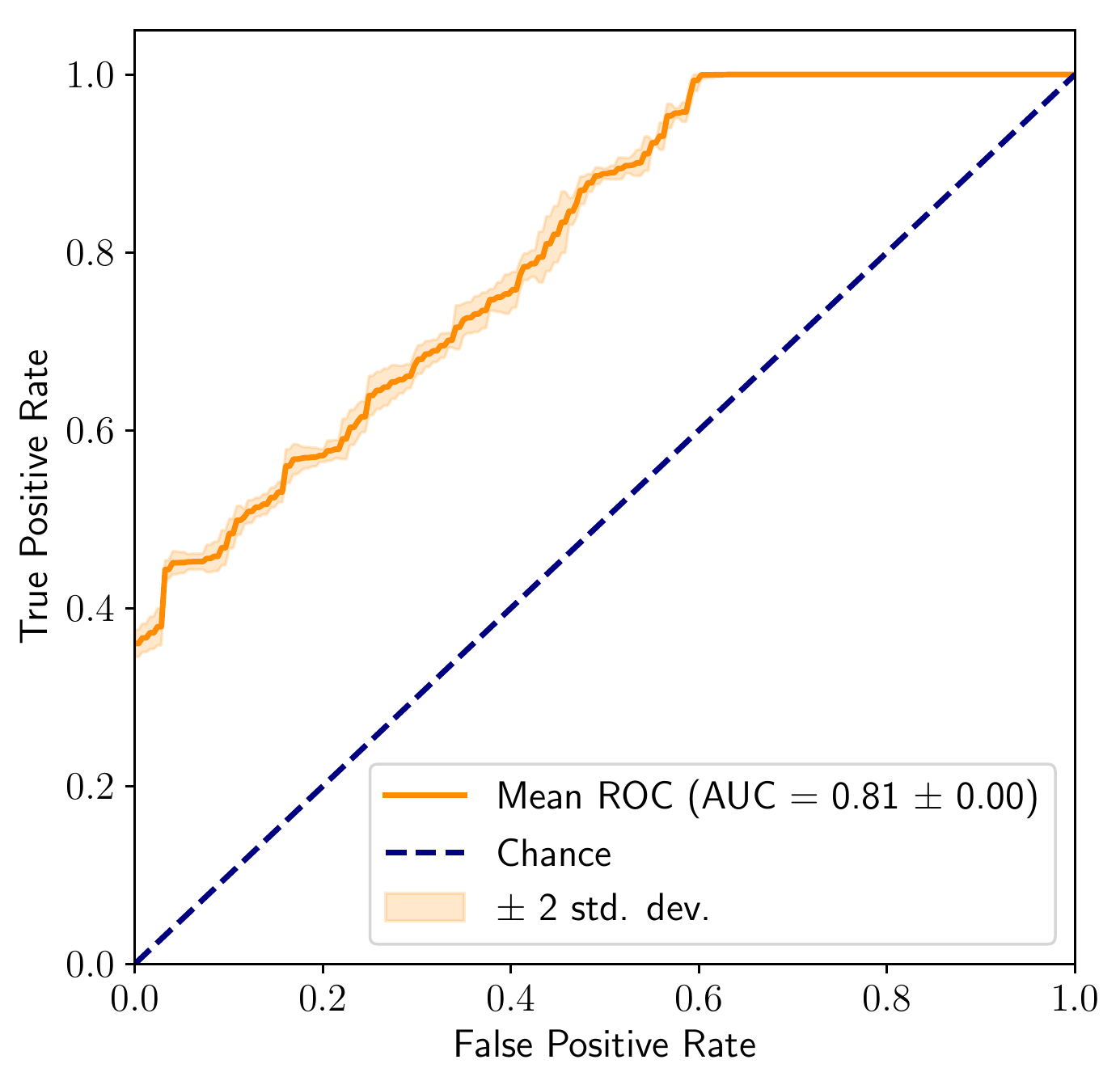}
        \caption{OMGP.\label{fig:roc_omgp}}
    \end{subfigure}
    \caption{Receiver-operating curves (ROC) for the mixtures of Gaussian process models. These ROC curves result from the experiments conducted in \cref{sec:gp_exp}. The mixture of Gaussian processes with BEEM inference is shown in \cref{fig:roc_beem} and the OMGP model's ROC curve is depicted in \cref{fig:roc_omgp}. As noted in \cref{sec:gp_exp} each model was applied to the same dataset (see \cref{fig:data_assoc}) 50 times, consequently the above mean ROC curve is shown plus/minus two standard deviations.\label{fig:roc_curves}}
\end{figure*}

In \cref{sec:gp_exp} a simple data-association experiment was conducted on the dataset shown in \cref{fig:data_assoc}, wherein the true number of clusters is $K=2$ and in total there are $N=250$ observations, each of which needs to be associated with a source generative process. From those experiments the receiver-operating curves are plotted in \cref{fig:roc_curves}.

From \cref{fig:roc_curves} a number of inferences can be drawn. First, the area-under-the ROC curve (AUROC) has been calculated and included with standard deviations over all experimental results. The AUROC of the data-association models corresponds to the probability that the model (MGPs w. BEEM or OMGP) will rank a randomly chosen positive example higher than a randomly chosen negative example. From this definition we see that the OMGP is, on average, better than the MGP w. BEEM at associating observations with the correct source. At the same time, we see from \cref{fig:roc_beem}, that taking the uncertainty bounds into account, yields comparable performance alongside OMGP. 

Now of course, the toy-dataset from whence the AUROCs were generated is simplistic, and there are many parameters to tune (see the next section). Certainly, it is possible to push the AUROC envelope further, for both models, by performing a hyper-parameter search using e.g. Bayesian optimisation, though that is outside the scope of this paper. This example demonstrates that MGP w. BEEM is capable of producing comparable performance to a much more complex model.

\subsection{More complex data-association experiment}

\begin{figure*}[ht!]
    \centering
    \begin{subfigure}[t]{0.49\textwidth}
        \centering
        \includegraphics[width = \columnwidth]{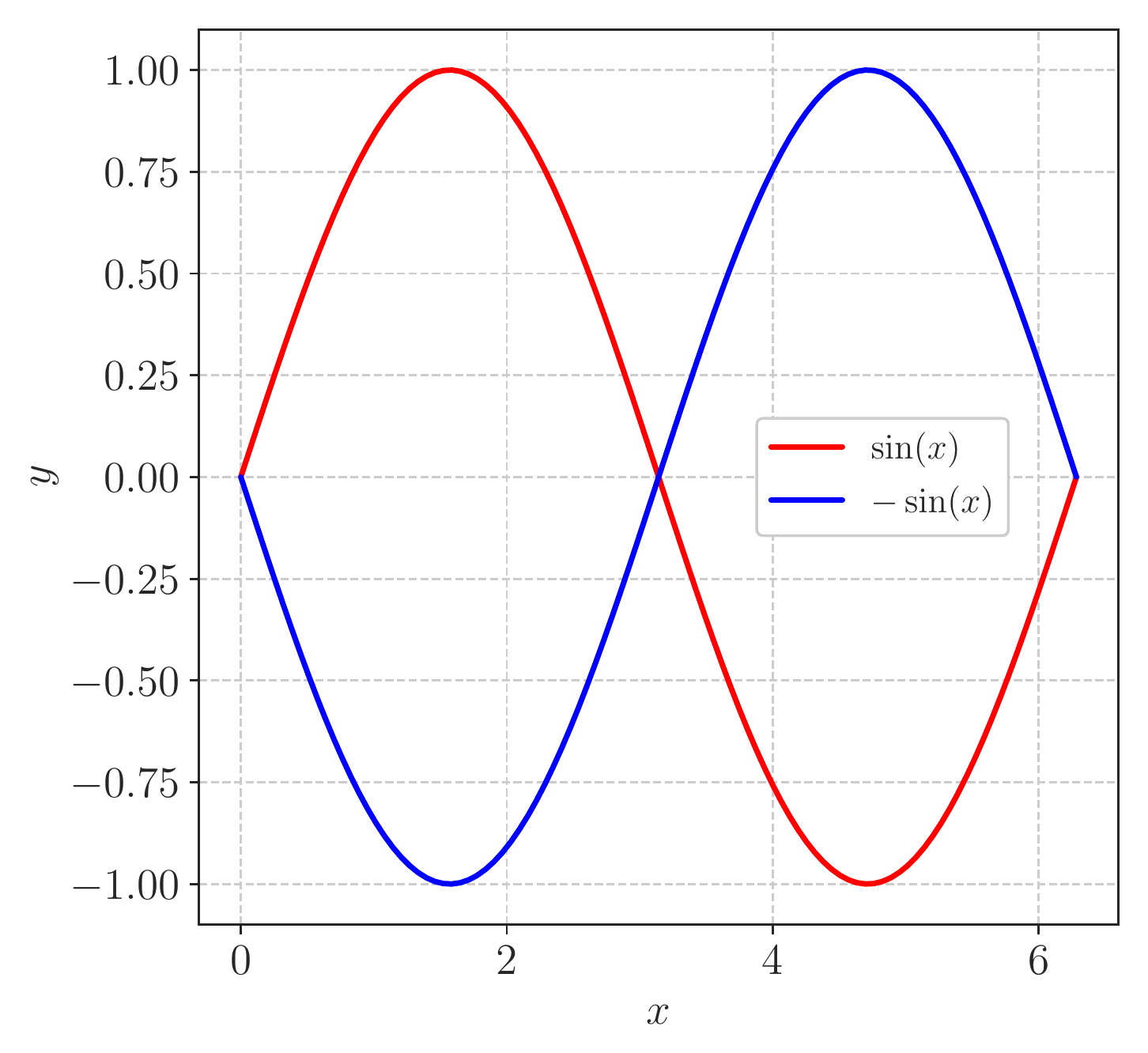}
        \caption{Two function representing the `true' underlying data generation processes.\label{fig:data_generating}}
    \end{subfigure}
    ~
    \begin{subfigure}[t]{0.49\textwidth}
        \centering
        \includegraphics[width = \columnwidth]{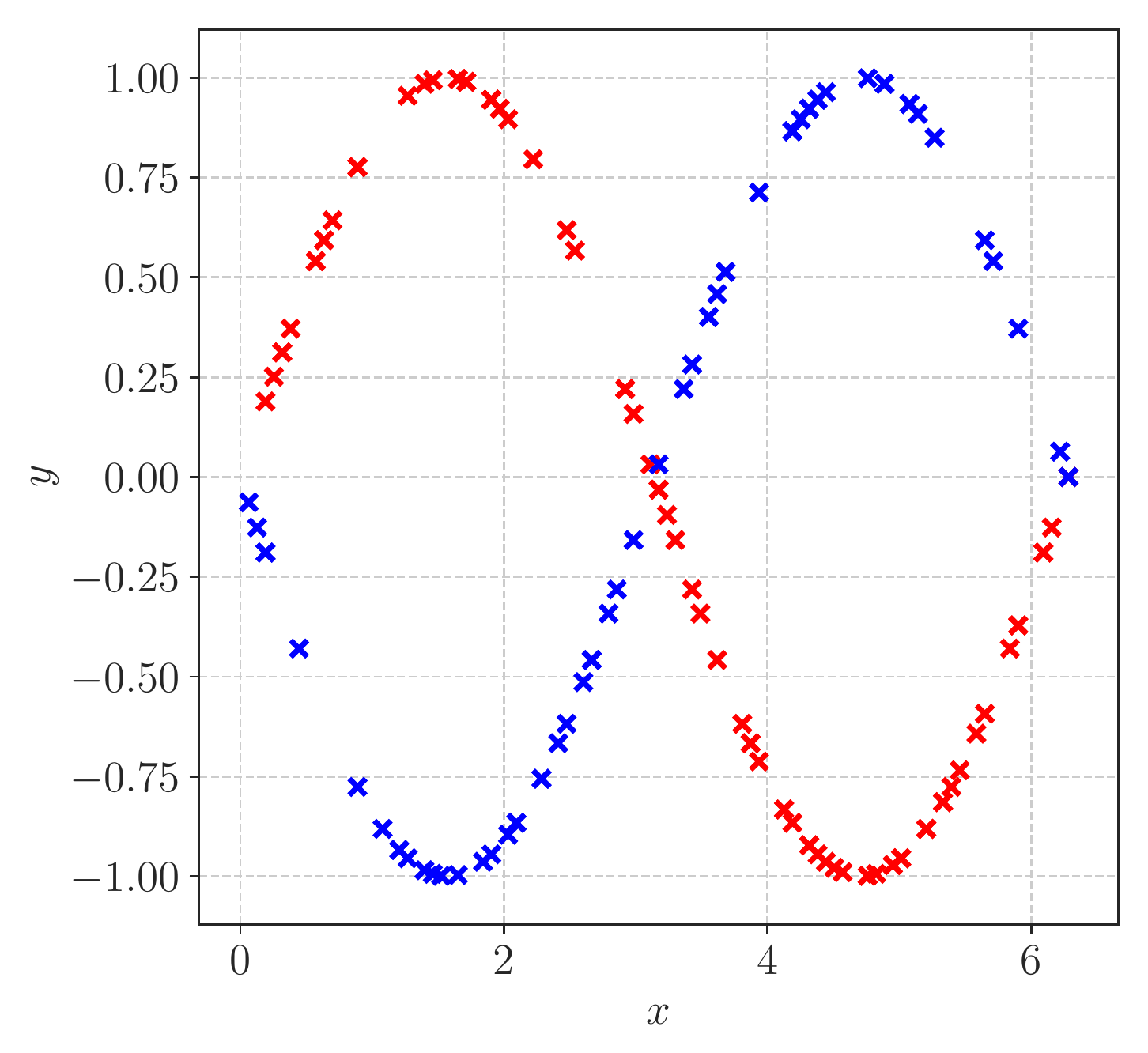}
        \caption{To create a more realistic dataset, data points are removed at random, resulting in sequences of \textcolor{red}{$N=75$} and \textcolor{blue}{$N=60$}. \label{fig:irregular_samples}} 
    \end{subfigure}
    \\
    \begin{subfigure}[t]{0.49\textwidth}
        \centering
        \includegraphics[width = \columnwidth]{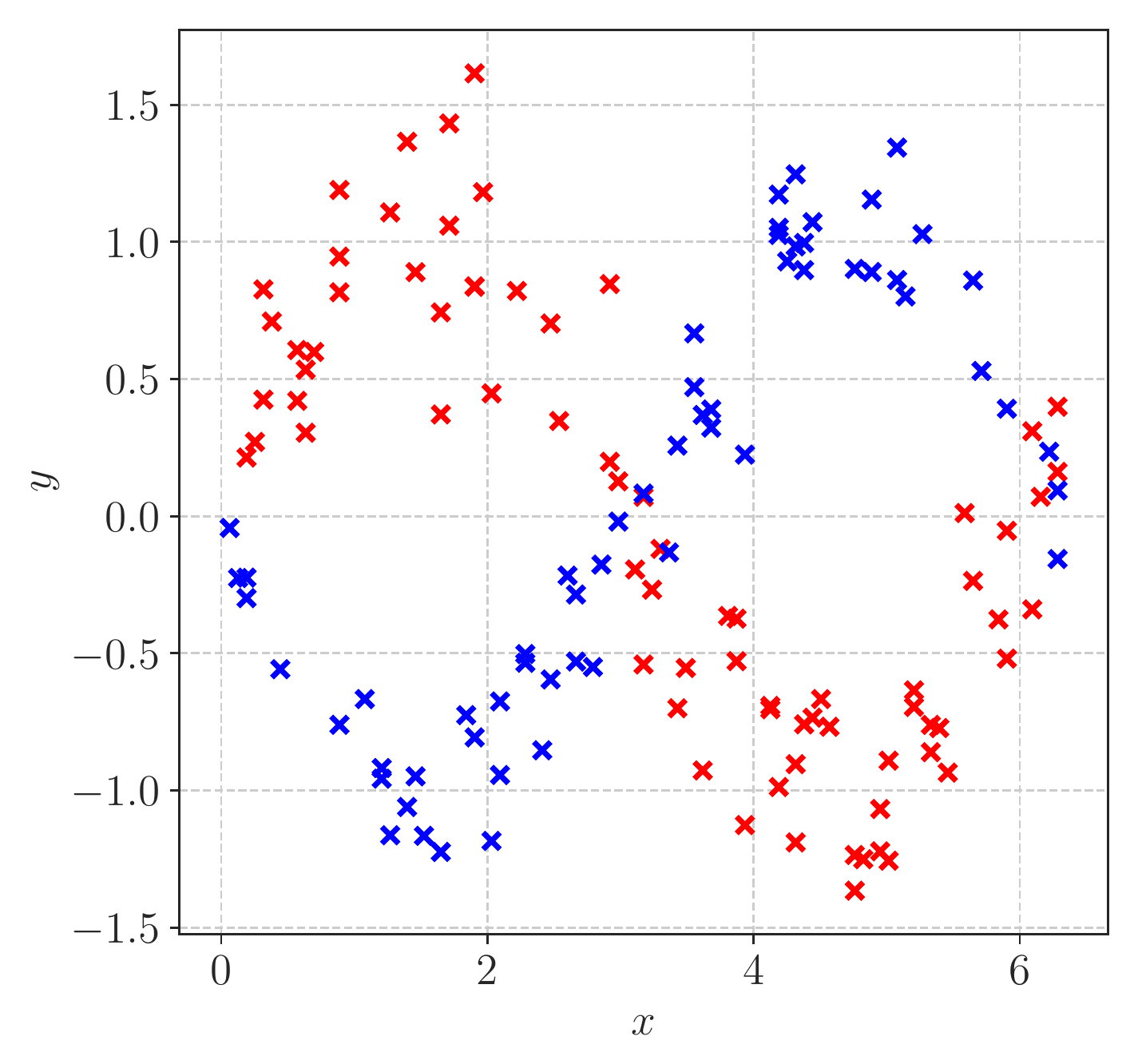}
        \caption{Two different additive white-noise processes are included: \textcolor{red}{$\mathcal{N}(0,\sigma^2=0.3)$} and \textcolor{blue}{$\mathcal{N}(0,\sigma^2=0.2)$}.\label{fig:noise_models}}
    \end{subfigure}%
    ~
    \begin{subfigure}[t]{0.475\textwidth}
        \centering
        \includegraphics[width = \columnwidth]{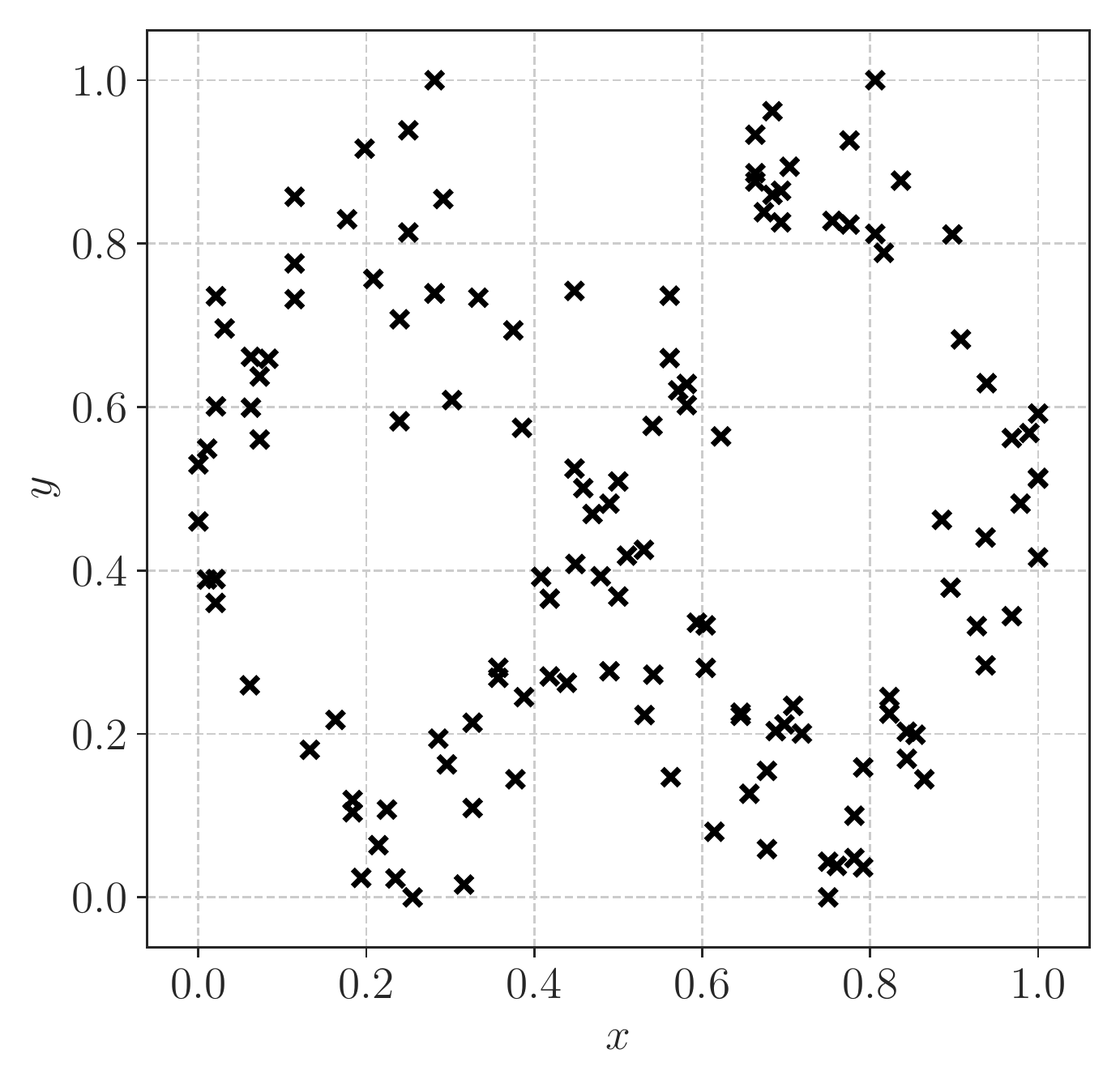}
        \caption{The final dataset as passed to the respective mixture models.\label{fig:data_final}}
    \end{subfigure}
    \caption{Synthetic data-generation process. The sub-figures within demonstrate the data-generation process, from top-left to bottom-right. This example is more complex than the one used in the body of this paper and has multiple overlapping areas. Colour-coded items in the sub-captions indicate a relation with the red and blue curves in the data-generating functions in \cref{fig:data_generating}. Moreover, the observation sets have been designed to have different noise-models \cref{fig:noise_models} and irregularly sampled \cref{fig:irregular_samples}. The normalised dataset, as seen by the data-association models is depicted in \cref{fig:data_final}.
    \label{fig:gp_ex_complex}
    }
\end{figure*}

In this section we conduct a more complex data-association experiment. It is inspired by the \citep[\S 4.1.1]{lazaro2012overlapping} but sacrifices dimensionality for a more complex noise regime, as well as allowing for a irregularly sampled observation space. Unlike the experiment conducted in \cref{sec:gp_exp} we provide a walk-through of the toy-data generation process in \cref{fig:gp_ex_complex}, this is to provide insight into what we want the MGP w. BEEM and the OMGP, to reproduce, as well as understand what layers of complexity need to be overcome, in order to approximate the original sources in \cref{fig:data_generating}. The original sources are chosen as a positive and a negative sinusoidal curve, each over one full cycle, and both have multiple regions of interaction (i.e. where they overlap).

\begin{figure*}[ht!]
    \centering
    \begin{subfigure}[t]{0.49\textwidth}
        \centering
        \includegraphics[width = \columnwidth]{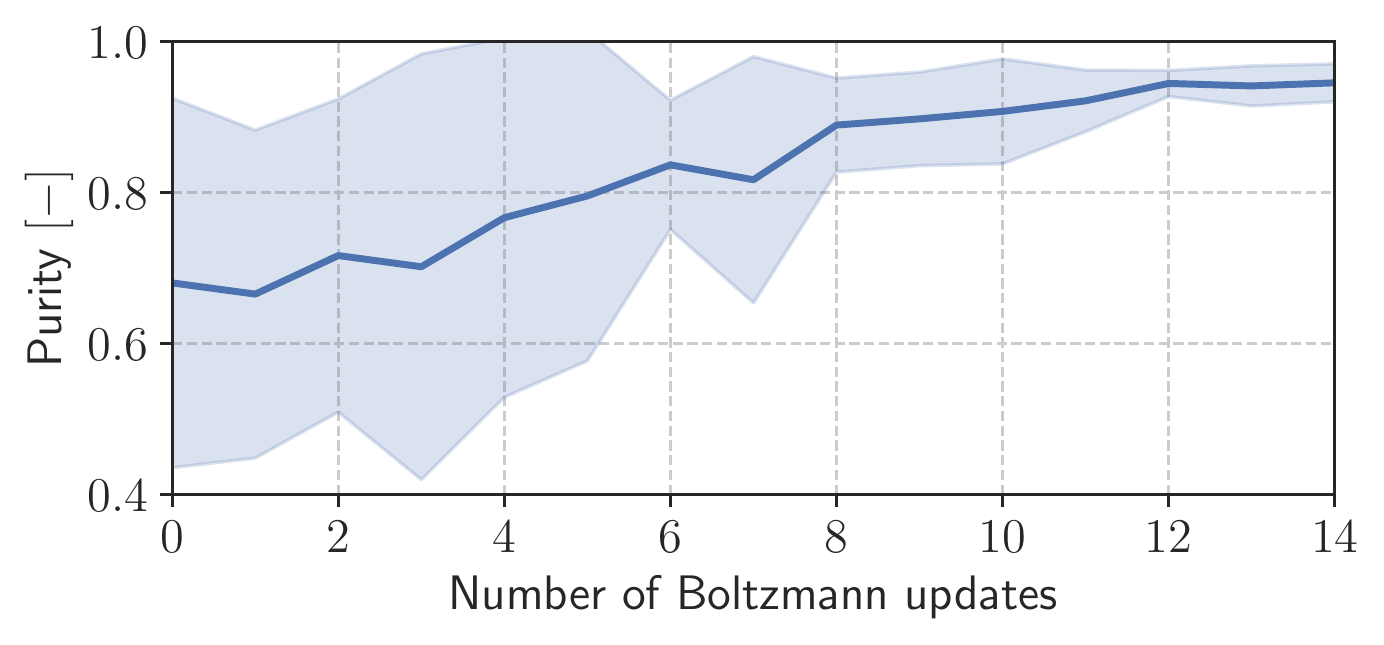}
        \caption{MGP w. BEEM --cluster purity.\label{fig:mgp_beem_purity}}
    \end{subfigure}%
    ~ 
    \begin{subfigure}[t]{0.49\textwidth}
        \centering
        \includegraphics[width = \columnwidth]{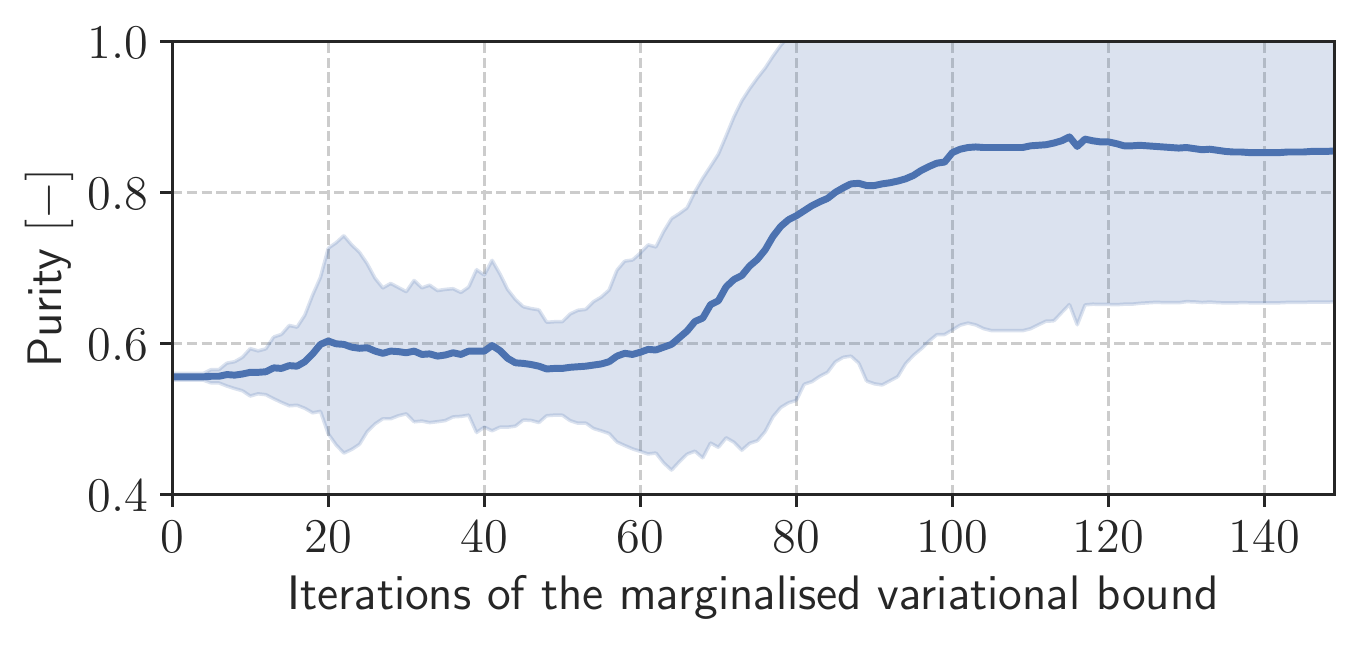}
        \caption{OMGP cluster purity $\sigma^2 = 0.1$.\label{fig:omgp_purity_01}}
    \end{subfigure}
    \par\bigskip 
    \begin{subfigure}[t]{0.49\textwidth}
        \centering
        \includegraphics[width = \columnwidth]{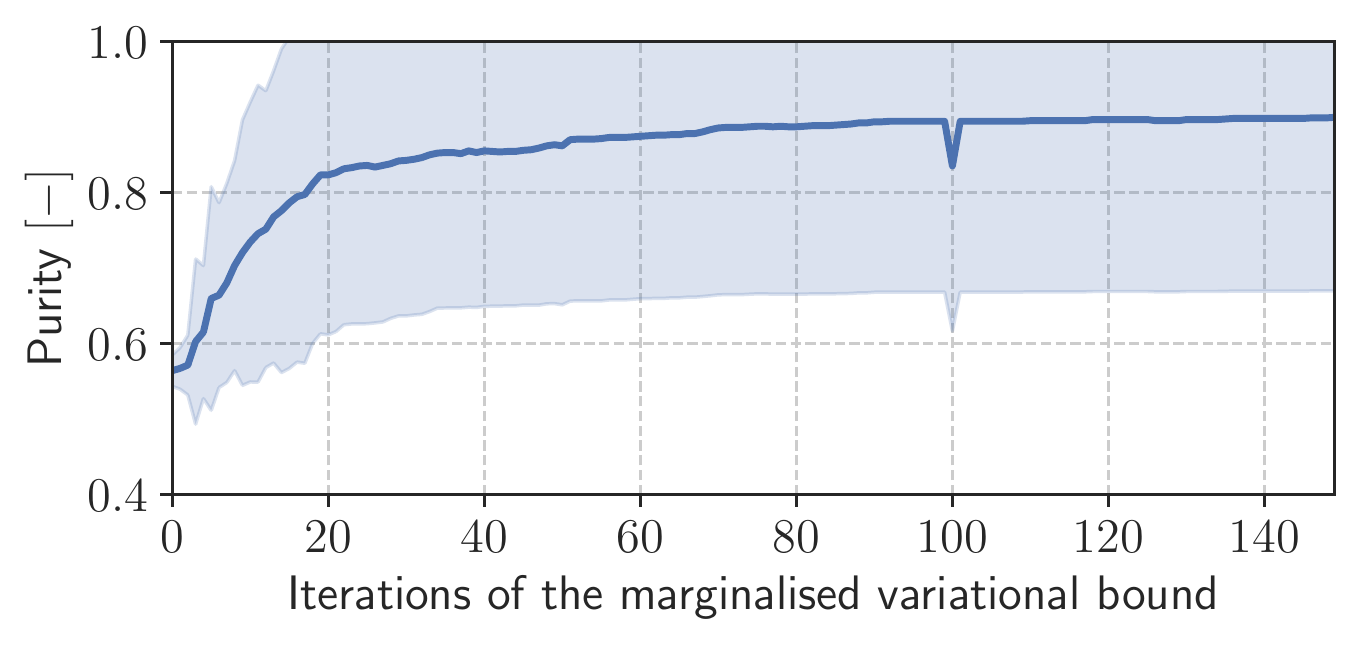}
        \caption{OMGP cluster purity with $\sigma^2 = 0.01$.\label{fig:omgp_purity_001}}
    \end{subfigure}
    \caption{Measured purity for each method. Purity is an external evaluation criterion of cluster quality, the percent of the total number of observations that were classified correctly, in the unit range $[0,1]$. Each method was run 10 times, the thick line in each plot shows the mean trend $\mu \pm 2\sigma$. Note that the horizontal axis is \emph{not} the same for both methods.\label{fig:gp_mix_difficult_results}}
\end{figure*}

As before, our implementation for the MGP used \texttt{GPy} \citep{gpy2014}, where each model was furnished with the model parameters found in \cref{tab:eight_params}. Both models are privy to the true number of clusters and were provided with the same periodic kernel -- for details on this kernel see \citep{Rasmussen:2005:GPM:1162254}. The output variance $\sigma^2$ determines the average distance of the function away from its mean, we note this because we performed two experiments with the OMGP where this parameter was varied (see \cref{tab:eight_params}). In total the OMGP's variational inference procedure was run for 150 iterations, and the GP base-models in BEEM were updated ten times each, for every Boltzmann update.  We again measure cluster quality based on \emph{purity} which is the percent of the total number of observations that were  correctly classified. Results are shown in \cref{fig:gp_mix_difficult_results}.

\begin{table}[!htbp]
    \centering
    \caption{Parameters used for both data-association models applied to the dataset described in \cref{fig:gp_ex_complex}.}
    \label{tab:eight_params}
    \def\arraystretch{1.3}%
    \begin{tabular}[t]{l|cc}
        \toprule 
        Parameter  & MGP w. BEEM & OMGP  \\
        \hline
        $K$ & $2$ & $2$ \\ 
        $\sigma^2$ & $0.1$ & $0.1$ \& $0.01$ \\ 
        Kernel & $\sigma^2\exp\left(-\frac{2\sin^2(\pi|x - x'|/p)}{\ell^2}\right)$ & $\sigma^2\exp\left(-\frac{2\sin^2(\pi|x - x'|/p)}{\ell^2}\right)$ \\ 
        Base model (GP) iterations & 10  & 150  \\ 
        Boltzmann updates & 15 & $-$  \\ 
        $\tau$ & $1.1$  &  $-$ \\ 
        $\alpha$ &  $0.97$ &   $-$\\ 
        $\varepsilon$ &  $1$ &   $-$\\ 
        \bottomrule 
    \end{tabular}
\end{table}

The results for the MGP w. BEEM are shown in \cref{fig:mgp_beem_purity}. Contrast this purity trend with those in \cref{fig:omgp_purity_01} and \cref{fig:omgp_purity_001}. The result in \cref{fig:mgp_beem_purity} displays an exploration-exploitation behaviour on part BEEM where the method initially has a large spread of cluster assignments, but as the temperature is reduced, it is also clear that a good point has been found on the likelihood surface. The OMGP on the other hand, has much lower purity spread initially, taken across all experimental runs, but increases the variance as the variational bounds are optimised \citep{lazaro2012overlapping}.
\FloatBarrier
\section{Related material -- expectation-maximisation initialisation}
\label{sec:EM_init_appendix}

This appendix covers related work on EM initialisation which, although not as relevant as the deterministic annealing EM approach, is still relevant in the grander scheme of finding good solutions due to EM.

Herein we compare and contrast methods for initialising the EM procedure. It is crucial that a good initialisation scheme is used for the EM algorithm, in order to find the MLE for the finite mixture model \citep{hu2015initializing}. Predominantly ours is an EM algorithm which, rather than heuristically or randomly allocating the initial clusters, uses an exploration-exploitation mechanism which searches the cluster-allocation space, with every new EM update. That being said, there are many other successful strategies for initialising the EM procedure. It should be noted that the initialisation task becomes harder as $K$ grows and the great majority of the following methods concern the GMM and little else. 

$K$-means clustering is by far the most common approach for initialising the EM algorithm \citep{hu2015initializing, Murphy2012, Barber2012}. There are drawbacks however, like the EM algorithm, the $K$-means method \emph{also} requires initialisation, which becomes more difficult as $K$ grows \citep{hu2015initializing}. Methods have been developed which better initialise the $K$-means method, such as $K$-means++ \citep{arthur2007k}. But, notes \citet{hu2015initializing}, $K$-means++ is still inadequate when $K\gg2$. The latter $K$-means derivative is one of a host of methods which seek to efficiently initialise the original \citep[\S 2.2.1]{hu2015initializing}. But in sum, they all suffer when $K$ becomes too large, and though there are plenty of studies which compare many of these $K$-means derivatives, there is no conclusive evidence that favours one method over another, consequently randomly initialising vanilla $K$-means remains the most widely used approach \citep{hu2015initializing}. That being said, \citet{blomer2013simple} found that $K$-means++ outperformed the comparison methods used in their study.


Because $K$-means itself is sensitive to initialisation, it is rendered somewhat undesirable for initialising the EM algorithm (as both are highly sensitive to the initialisation). Suitable alternatives are methods which do not require initialisation themselves (like vanilla $K$-means), one such method is hierarchical clustering \citep{hu2015initializing}. An advanced form of hierarchical clustering was used for initialising EM and was described by \citet{fraley1998algorithms}, for fitting finite mixture models \citep{hu2015initializing}. 

\citet[\S 2.4]{hu2015initializing} notes an interesting aside w.r.t. EM initialisation, wherein authors have modified the EM algorithm itself to be robust \emph{against} bad initialisation. But, notes \citet{hu2015initializing}, these approaches are not strictly speaking EM, but have strong connections to the initial algorithm, first proposed by \citep{dempster1977maximum}. Their primary purpose is to improve the clustering results when the GMM is employed \citep{hu2015initializing}.

To conclude this section; there is no way to determine \emph{the} best initialisation algorithm, which universally and consistently, achieves the best performance in all application domains. That performance will always depend on the quality and size of the data, as well as the allowed computational cost \citep{blomer2013simple}.




\end{document}